\documentclass[onecolumn]{article}

\usepackage{PRIMEarxiv}

\usepackage{graphicx}  
\usepackage{amsfonts}
\usepackage{amsmath}
\usepackage{amsmath}
\usepackage{algorithm}      
\usepackage{algpseudocode}  
\usepackage{booktabs}
\usepackage{siunitx}
\usepackage[table, svgnames, dvipsnames]{xcolor}
\usepackage{float}
\usepackage{accents}
\usepackage{tabularx}
\usepackage{array}
\usepackage{comment}
\usepackage{mathtools}
\usepackage{hyperref}
\usepackage{wrapfig}
\usepackage{picinpar}
\usepackage{afterpage}
\usepackage[numbers]{natbib} 
\DeclareMathOperator*{\argmin}{arg\,min}

\DeclareMathOperator{\EX}{\mathbb{E}}
\usepackage{abstract}

\usepackage{hyperref}

\definecolor{light-gray}{gray}{0.95}
\definecolor{Cerulean}{HTML}{00A2E3}
\definecolor{pinegreen}{HTML}{009B55}
\colorlet{LightCerulean}{Cerulean!30}

\definecolor{SeaGreen}{HTML}{3FBC9D}
\colorlet{SeaGreen}{SeaGreen!90}

\newcolumntype{P}[1]{>{\centering\arraybackslash}p{#1}}

\newcommand\Algphase[1]{%
\vspace*{-.7\baselineskip}\Statex\hspace*{\dimexpr-\algorithmicindent-2pt\relax}\rule{\textwidth}{0.4pt}%
\Statex\hspace*{-\algorithmicindent}\textbf{#1}%
\vspace*{-.7\baselineskip}\Statex\hspace*{\dimexpr-\algorithmicindent-2pt\relax}\rule{\textwidth}{0.4pt}%
}

\newcommand*{\belowrulesepcolor}[1]{%
  \noalign{%
    \kern-\belowrulesep 
    \begingroup 
      \color{#1}%
      \hrule height\belowrulesep 
    \endgroup 
  }%
} 
\newcommand*{\aboverulesepcolor}[1]{%
  \noalign{%
    \begingroup 
      \color{#1}%
      \hrule height\aboverulesep 
    \endgroup 
    \kern-\aboverulesep 
  }%
}

\usepackage{fancyhdr}

\makeatletter

\title{Ricci flow regularization in latent spaces for the forward learning of partial differential equations}
\author{Andrew Gracyk\footnotemark[1]}
\date{}

\begin{document}

\vspace{-20mm}
\begin{@twocolumnfalse}
\maketitle
\vspace{-6mm}
\begin{abstract}
\vspace{0mm}
We present a manifold-based machine learning encoder-decoder method for learning dynamics in time, notably partial differential equations (PDEs), in which the manifold latent space evolves according to Ricci flow. This can be accomplished by parameterizing the latent manifold stage and subsequently simulating Ricci flow in a physics-informed setting, matching manifold quantities so that Ricci flow is empirically achieved. We emphasize dynamics that admit low-dimensional representations. With our method, the manifold, induced by the metric, is discerned through the training procedure, while the latent evolution due to Ricci flow provides an accommodating representation. By use of this flow, we sustain a canonical manifold latent representation for all values in the ambient PDE time interval continuum. We showcase that the Ricci flow facilitates qualities such as learning for out-of-distribution data and adversarial robustness on select PDE data. Moreover, we provide a thorough expansion of our methods in regard to special cases which allow higher-dimensional representations, such as Ricci flow on the hypersphere and neural discovery of non-parametric geometric flows with entropic strategies.
\vspace{4mm}
\end{abstract}
\vspace{2mm}
\end{@twocolumnfalse}

\footnotetext[1]{University of Illinois Urbana-Champaign, supported from DIGIMAT with NSF Grant No. 1922758}

\section{Introduction}

Data-driven techniques fueled by machine learning act as a meaningful lens for exploring numerical-based approaches and solutions to partial differential equations (PDEs) \citep{Lu_2021} \citep{li2021fourier} \citep{wang2021learning} \citep{ovadia2023realtime} \citep{cao2023lno} \citep{takamoto2023learning} \citep{dool2023efficient}
\citep{goswami2022physicsinformed}. We will study the discrete autoencoder paradigm \citep{lopez2022gdvaes} \citep{glyndavies2023phidvae} \citep{NEURIPS2022_6d5e0357} \citep{huang2022metaautodecoder} for ambient time-dependent dynamics, as well as the auxiliary outcomes of latent structure due to Ricci flow specifically in learning strategies for such dynamics. Encoder-decoder methods are highlighted by a composition of neural networks with an intermediary stage, typically unrestricted without additional regularization or penalization. Encoder-decoder methods alone serve as a baseline method with limited properties, but by leveraging dynamic or geometric properties, they gain a general performance boost along with interpretability, extrapolation-ability, and robustness \cite{lopez2022gdvaes} \cite{lopez2021variationalautoencoderslearningnonlinear} \cite{chadebec2022geometricperspectivevariationalautoencoders} \cite{syrota2024decoderensemblinglearnedlatent} \cite{shukla2019geometrydeepgenerativemodels} \cite{jang2023geometrically} \cite{khan2024adversarialrobustnessvaeslens} \cite{10.1007/978-3-030-58580-8_2} \cite{Tron2024}, possibly with curvature-related regularization effects. Geometric dynamic variational autoencoders (GD-VAEs) \cite{lopez2022gdvaes} have shown great promise in learning discrerized time-dependent dynamics, which are highlighted by an underlying latent space with geometric, dynamic features or dynamics in the latent space itself. We focus on an encoder-decoder method, emphasizing a low-dimensional manifold latent space which is imposed with dynamical evolution, that also holds the advantages that GD-VAEs achieve, including competitive error, generalization, extrapolation ability, and robustness. Even moreso, with our method, the dynamics in the manifold itself provide an accommodating manifold representation over the entire ambient PDE's time interval, and the ambient PDE's progression is learned in a larger scale of time by corresponding to a continuum.

\vspace{2mm}

A geometric flow, the foundation of our method, is a tool to perform dynamics among the manifold, in which we choose Ricci flow \citep{toppingricciflow} \citep{ricciflowsheridan} \citep{calegariricciflow}. We first represent the initial PDE data in a domain to parameterize the manifold, which is then encoded directly to a point on the manifold. The points that compose this manifold evolve according to such a flow, in which we use a physics-informed neural network to simulate the evolution, matching quantities among the manifold to those that solve the physics-informed flow.

\vspace{2mm}

Ricci flow offers several advantages. It is an example of an intrinsic geometric flow, meaning the relevant quantities computed in the PDE flow are inherent to the manifold itself \citep{taftgeoflows}, and so the manner in which it is embedded in space is not involved in the PDE computation and is otherwise irrelevant. Our method, captured by an intrinsic flow, is highlighted by this fact. This allows the manifold to displace through space, aside from the movements due to the flow, during its period of evolution. The embedding is irrelevant, and can change to maximize effectiveness in solving the objective function. This displacement, including rotation, corresponds to a vector field, causing movement. Furthermore, simulation of an extrinsic flow is nontrivial, as it would typically require construction of vectors orthogonal to the tangent space, and each manifold particle would need to follow this trajectory. This would likely be done through the use of neural ordinary differential equations. Furthermore, Ricci flow exhibits curvature uniformization properties, which deform the volume element as time progresses. This is notable because the learned representations can be done through both embedding displacement and manifold evolution, allowing greater flexibility in latent expressions. Lastly, there are not many well-known intrinsic geometric flows throughout classical literature. For example, mean curvature flow is well-known, but this is extrinsic. Calabi flow is well-known, but this is for complex manifolds. Yamabe flow is more consistent with our special cases.

\vspace{2mm}

The intermediary stage of the parameterization domain acts similarly to the unstructured data in Euclidean space in the GD-VAE setting in the sense that it is fixed and with less structure than before its mapping to the manifold, except our parameterization domain is of lower dimension, being the intrinsic dimension, than the embedding dimension. Such a parameterization contains lesser structure than the embedding latent stage. The mapping from unstructured data to the manifold for GD-VAEs acts as a way to progress the encoding from data to that of a later state in the ambient data evolution.

\vspace{2mm}

We compare our methods to baselines. In one such baseline, we provide no restrictions on the latent space, which we let develop naturally in training. As demonstrated in \cite{chadebec2022geometricperspectivevariationalautoencoders} \cite{davidson2022hypersphericalvariationalautoencoders}, it is possible Riemannian structures develop in the respective latent spaces on their own accord in a learning task with no concrete constraints. In this sense, we investigate the empirical performance of Ricci flow specifically, and so we provide empirical support of our methods in comparison to these unrestricted baselines, whether or not geometric properties formulate independently in these spaces without any enforced properties of our choosing.

\vspace{2mm}

\begin{figure}
  \vspace{0mm}
  \centering
  \includegraphics[scale=0.7]{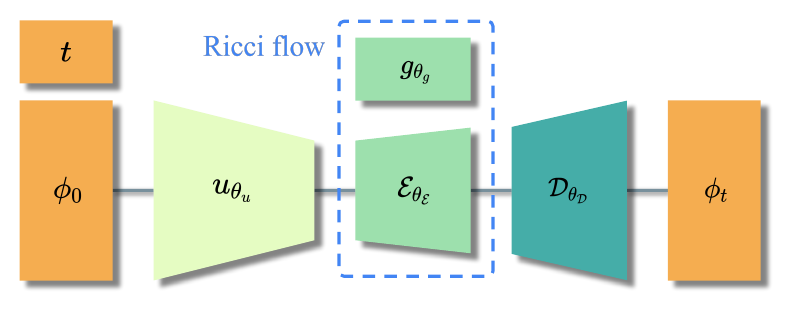}
  \caption{We illustrate our method: we construct a series of neural networks with a manifold latent space. An intermediary Riemannian metric neural network is simulated with Ricci flow, in which we match the inner products of the tangents with the learned manifold with such a metric.}
  \label{fig:ricci_flow_diagram}
\end{figure}

While the implicitly-learned latent space captured by an intrinsic flow allows high precision in the task of replicating data from the same underlying distribution as seen as part of the training procedure, it is of interest to achieve precision in areas such as generalization of the data and extrapolation of data with fundamental qualities of those seen in training, but sufficiently different to be categorized as data that is extended outside of the training procedure. It is also of interest to develop methods that are highly robust against adversarial examples of test data, such as those corrupted with noise, given networks trained on dynamics of the original functions to be learned \citep{raonić2023convolutional} \citep{Thanasutives_2023}
\citep{adesoji2022evaluating}. Many existing paradigms emphasize high performance on in-distribution data \citep{melchers2024neural} \citep{ovadia2023realtime} \citep{adesoji2022evaluating}.The manifold latent space with dynamics offers promise in extending data-based methods for learning PDEs from a baseline to one that not only produces high in-distribution performance, but also yields greater ability to infer dynamics from never-before-seen data, enhance robustness, as well as incorporate features of the data into such a structure \citep{lopez2022gdvaes}, cultivating interpretability.

\section{Problem formulation}

\textbf{Problem setting.} The task of interest is to learn the solution of a parameterized partial differential equation of the form $ \partial_t \phi + \mathcal{D}_{\alpha}[\phi] = 0 , $
subject to initial data $\phi(\cdot, 0) = \phi_0$, where $\phi : \mathcal{X} \times [0,T] \rightarrow \mathbb{R}$ is the solution, and $\mathcal{D}_{\alpha}$ is a differential operator with parameter $\alpha$. $\alpha$ may depend on evaluations over the input space $\mathcal{X}$ as well as time $t \in [0,T]$. In particular, we seek a mapping
\begin{gather}
\Gamma : (\phi_0, t) \in (C \cap W^{k,2})(\mathcal{X}; \mathbb{R})  \times [0,T] \rightarrow \phi_t \in (C \cap W^{k,2})(\mathcal{X}; \mathbb{R}) , \ \text{or discretized,} 
\\
\Gamma^{\dagger} :  \{ \tilde{\phi}_0 \in \mathbb{R}^N : \tilde{\phi}_0 = \phi_0 |_{\Omega}, \phi_0 \in (C \cap W^{k,2})(\mathcal{X}; \mathbb{R})  \} \times [0,T]  \rightarrow \{ \tilde{\phi}_t \in \mathbb{R}^N : \tilde{\phi}_t = \phi_t |_{\Omega}, \phi_t \in (C \cap W^{k,2})(\mathcal{X}; \mathbb{R}) \} ,
\end{gather}
for mesh $\Omega$ such that the continuum version $\Gamma$ above is discretized, as is below with $\Gamma^{\dagger}$.
We learn this over numerous instances of data. We devise an autoencoder, characterized by a latent space that consists of collections of points that lie directly along a continuum of manifolds $\mathcal{M}_{\hat{\tau}}$ embedded in Euclidean space $\mathbb{R}^d$, $\hat{\tau} \in [0,\tau]$ for some $\tau$, to learn $\tilde{\phi}_t$.

\vspace{2mm}

\textbf{Ricci flow.} Ricci flow is a natural framework for accomplishing goals of incorporating input data features into a manifold structure with generalization capability while allowing high precision in the task of producing the output. Ricci flow unites the separate features of an evolving latent space and a manifold latent space into a single scheme. Ricci flow is a partial diffferential equation that describes changes in the Riemannnian metric. Let $g_{ij} : \mathcal{U} \times [0,\tau] \rightarrow \mathbb{R}$ be the components of the Riemmannian metric matrix $g$ over the manifold parameterization domain (local coordinates) $\mathcal{U}$ and a time domain $[0,\tau]$. Ricci flow is the metric evolution equation \citep{toppingricciflow} \citep{ricciflowsheridan}
\begin{equation}
\label{Ricci_flow}
\partial_t g(u, t) = -2 \text{Ric} (g(u,t)) ,
\end{equation}
for $u \in \mathcal{U}, t \in [0,\tau]$, such that the metric solved by \ref{Ricci_flow} reflects that of the manifold $\mathcal{M}_{\hat{\tau}}$ over its respective time interval. The solution to this PDE can be regarded as an $m \times m$ matrix, $g(u,t)$, with components $g_{ij}$. The right-hand term $\text{Ric} (g(u,t))$ is the Ricci tensor. Ricci flow is similar to a geometric diffusion equation in which curvature is uniformized, and geometries contract or expand.

\vspace{2mm}

\textbf{Metric coefficients.} The coefficients of the Riemannian metric, $g_{ij}$, can be computed using the tangent vectors to the manifold at a location corresponding to a point in the manifold parameterization domain local coordinates and time domain $\mathcal{U} \times [0,\tau]$. The induced metric and its coefficients can be computed at general times as \citep{elemofdiffgeo}
\begin{equation}
\label{eqn:metric_coefficient}
0 \preceq g = {\sum \sum}_{ij} g_{ij} du^i \otimes du^j \in \Gamma(\text{Sym}^2 T^* \mathcal{M}), \ \ \ \ \ g_{ij}(u,t) \approx \langle \partial_i \mathcal{E}(u,t), \partial_j \mathcal{E} (u,t) \rangle ,
\end{equation}
where $\partial_{\ell}$ is the component-wise partial derivative with respect to the $\ell$-index in $u = (u^1, u^2, \hdots, u^m)$ belonging to the parameterization domain $ \mathcal{U}$, and $\mathcal{E} : \mathcal{U} \times [0,\tau] \rightarrow \mathbb{R}^d$ is the function mapping to the manifold continuum. We will hold $m = d-1$. Here, $\langle \cdot, \cdot \rangle$ denotes the standard inner product. The Ricci flow PDE is primarily governed based on initial coefficients $g_{ij,0}$, but in our methodology, it of interest to compute the metric coefficients among arbitrary times $\tilde{\tau}_i \in U([0,\tau])$ in a uniform collocation procedure. The behavior of the manifold is governed by the change of the metric coefficients under Ricci flow such that equation \ref{eqn:metric_coefficient} holds.

\vspace{2mm}

\textbf{Ricci tensor.} Given explicit knowledge of the Riemannian metric coefficients, one may compute the Ricci tensor using Christoffel symbols and the Riemannian curvature tensor. We have the collections of the Christoffel symbols, $\{{\Gamma_{ij}}^l\}_{i,j,l}, 1 \leq i,j,l \leq m$ evaluated at particular $(u,t)$ \citep{elemofdiffgeo}, the Riemannian tensor $\{ {R}_{i}{}^{l}{}_{jk} \}_{i,l,j,k}, 1\leq i,l,j,k \leq m$ \citep{elemofdiffgeo}, and the Ricci tensor $\{ \text{Ric}_{ik} \}_{i,k}, 1 \leq i,k \leq m$ \citep{calegariricciflow}, given by 
\begin{align}
& {\Gamma_{ij}}^l = \frac{1}{2}  g^{kl} ( \partial_j g_{ik} - \partial_k g_{ij} + \partial_i g_{kj} ) ,
\\
& {{R}_i}{}^{l}{}_{jk} = \partial_j {\Gamma_{ik}}^l - \partial_k {\Gamma_{ij}}^l  +  {{\Gamma}_{ik}}^p {\Gamma_{pj}}^l - {\Gamma_{ij}}^p {\Gamma_{pk}}^l  ,
\\
\label{eqn:ricci_tensor}
& \text{Ric}_{ik} = \text{Ric}(g(u,t))_{ik}   =  {{R}_i}{}^{l}{}_{lk}  .
\end{align}
We use Einstein notation, and so repeated indices in a term are summed. $g^{kl}$ denote the elements of the inverse matrix of $g$ at such a $(u,t)$. The Ricci tensor is a contraction of the Riemannian tensor, which is equivalently an analog of the trace of the Riemannian tensor. The above gives the corresponding Ricci tensor at a particular $(u,t)$.

\section{Methods}
\label{sec:methods}

Ultimately, we seek a series of functions such that
\begin{equation}
\tilde{\phi}_t \approx  ( \mathcal{D}\circ \mathcal{E} \circ u) ( \tilde{\phi}_0, t)  ,
\end{equation}
for approximate PDE data $\tilde{\phi}$, decoder $\mathcal{D}$, encoder $\mathcal{E}$, and parameterization $u$ of local coordinates. The manifold, embedded in a low-dimensional space while subject to Ricci flow, is placed upon the encoding $\mathcal{E}$.

\vspace{2mm}

\textbf{Neural network design.} PDE solution information acts as input training data into a parameterization neural network $u_{\theta_{u}}$, mapping such data to a point in the parameterization domain $\mathcal{U}$ of the manifold continuum $\mathcal{M}_{\hat{\tau}}$. In particular, we construct a neural network $u : \mathbb{R}^N \times \Theta_{u} \rightarrow \mathcal{U}$, where data input is an $N$-dimensional vector containing data of the PDE initial condition disccretization $\{ \tilde{\phi}_0^i \}_{i}$, $\tilde{\phi}_0^i = (\phi^i(x_1, 0), \hdots, \phi^i(x_N,0))$, where $x_j$ is in mesh $\Omega$, and $i$ is an index among the collection of training data. We denote $u_{\theta_{u}}(\tilde{\phi}) = u(\tilde{\phi},\theta_{u})$. The purpose of this neural network, instead of directly mapping PDE information into a manifold latent space, is so that the partial derivatives $\partial_i \mathcal{E}(u,t)$, hence tangent vectors, of the manifold may be taken with respect to the same domain in which the metric $g$ is defined. Without this, \ref{eqn:metric_coefficient} cannot be formulated, and the manifold cannot be learned in training.

\vspace{2mm}

We propose a physics-informed neural network (PINN) framework for finding the metric solution $g$ to the Ricci flow equation. The PINN $g_{\theta_g}$ is constructed via standard neural network architecture, but is substituted into the pertinent PDE and is subsequently minimized as a PDE residual, as this is set to $0$ in what should be satisfied \citep{raissi2017physics} \citep{wang2023experts}
\citep{hao2023physicsinformed}. Differentiation of the PINN can be performed using automatic differentiation techniques found in deep learning libraries. To construct the metric, let $g : \mathcal{U} \times [0,\tau] \times \Theta_g \rightarrow \mathbb{M}^{m \times m}$, where $\mathbb{M}^{m \times m}$ is the set of $m \times m$ matrices. Since the metric should satisfy Ricci flow, the PDE residual to be minimized in our PINN setup is of the form \citep{jain2022physicsinformed}
\begin{equation}
\Big|\Big| \ \Big|\Big| \partial_t g_{\theta_g} + 2 \text{Ric}(g_{\theta_g}) \Big|\Big|_F^2 \ \Big|\Big|_{L^1(\mathcal{U} \times [0,\tau] ) } \propto \int_{0}^{\tau} \int_{\mathcal{U}} \sum_{ij} ( \partial_t g_{\theta_g,ij}(u,t) + 2\text{Ric}(g_{\theta_g}(u,t))_{ij} )^2 \rho(u) du dt,
\end{equation}
with an appropriate measure $\mu$ absolutely continuous with respect to Lebesgue measure on $\mathcal{U}$ to weight the domain such that the loss is uniform with respect to the input data, i.e. the physics loss is not uniform with respect to $\mathcal{U}$: $\rho$ is the corresponding Radon-Nikodym derivative. In practice, the physics loss is evaluated discretely with respect to the input data, so $\rho$ is a continuous interpolating extension to model the density of these points. Since the training loss runs over initial PDE data $\tilde{\phi}_0$, we will develop an analogous loss function with respect to this data.

\vspace{2mm}

Lastly, we construct encoder and decoder networks. The encoder maps a point along the learned parameterization domain directly onto the manifold embedded in Euclidean space at a specified time, and so the encoder additionally has a temporal input parameter corresponding to a state of evolution of the manifold. The encoder is given by $\mathcal{E} : \mathcal{U} \times [0,\tau] \times \Theta_{\mathcal{E}} \rightarrow \mathcal{M}_{\hat{\tau}}$. The decoder, $\mathcal{D} : \mathcal{M}_{\hat{\tau}} \times \Theta_{\mathcal{D}} \rightarrow \mathbb{R}^N$, maps the manifold point in Euclidean coordinates directly to known PDE solutions $\{ \tilde{\phi}_t^i \}_i$ with training data given by
\begin{equation}
\bigcup_{i} \bigcup_{t \in \{ j \Delta t, j \in \mathbb{N}, \Delta t \in \mathbb{R}^+ \} } \Bigg\{  \tilde{\phi}_t^i = (\phi^i(x_1,t), \hdots, \phi^i(x_N,t)) : \tilde{\phi}_t^i = \phi_t^i |_{\Omega}, \phi_t^i \in \mathcal{C}(\mathcal{X}; \mathbb{R}), x_j \in \Omega, 1 \leq j \leq N \Bigg\} ,
\end{equation}
but the framework can also be evaluated arbitrarily with respect to time if desired. The index $i$ runs over numerous PDE solutions, and $\Delta t$ is a time increment. A series of compositions of functions is taken to yield the final autoencoder output, with a loss restriction imposed at the intermediate step to ensure the desired evolution of Ricci flow is satisfied.

\vspace{2mm}

\textbf{Loss function.}  We construct our training objective function as follows:
\begin{align}
\begin{cases}
& \theta^* =  \argmin_{\theta \in \Theta} \text{Objective} = \mathcal{L}_{Ric}(\theta) + \lambda_{dec} \mathcal{L}_{dec}(\theta) + \lambda_{met} \mathcal{L}_{met}(\theta) \\
& \text{Objective} = \EX_{t \sim U[0,T]} \EX_{\phi_0 \sim \Phi}  \Big[ \frac{1}{|\mathcal{U}|^2} || \partial_t g_{\theta_g} ( u, \tilde{\tau}) + 2 \text{Ric}(g_{\theta_g}(u, \tilde{\tau})) ||_F^2
\\ &  + \lambda_{dec}  \frac{1}{N} || \mathcal{D}_{\theta_{\mathcal{D}}}(\mathcal{E}_{\theta_{\mathcal{E}}}(u ,\hat{\tau}) ) - \tilde{\phi}_t  ||_2^2    + \lambda_{met}  \frac{1}{|\mathcal{U}|^2} || g_{\theta_g}(u, \tilde{\tau}) - (J \mathcal{E}_{\theta_{\mathcal{E}}})^TJ \mathcal{E}_{\theta_{\mathcal{E}}}   ||_F^2 \Big] . 
\end{cases}
\end{align}
We use notation $u = u_{\theta_u} (\tilde{\phi}_0)$ and $J\mathcal{E}_{\theta_{\mathcal{E}}}$ to denote the Jacobian of vector $\mathcal{E}_{\theta_{\mathcal{E}}}$ with respect to $u$, i.e. $(J \mathcal{E}_{\theta_{\mathcal{E}}}(u, \tilde{\tau}))^TJ \mathcal{E}_{\theta_{\mathcal{E}}}(u, \tilde{\tau}) = (J \mathcal{E}_{\theta_{\mathcal{E}}})^T J \mathcal{E}_{\theta_{\mathcal{E}}}$ is the matrix of inner products
\begin{align}
& (J \mathcal{E}_{\theta_{\mathcal{E}}})^T J \mathcal{E}_{\theta_{\mathcal{E}}}  = 
\begin{pmatrix}
|| \partial_1 \mathcal{E}_{\theta_{\mathcal{E}}} (u, \tilde{\tau}) ||_2^2 & \langle \partial_1 \mathcal{E}_{\theta_{\mathcal{E}}}(u, \tilde{\tau}), \partial_2\mathcal{E}_{\theta_{\mathcal{E}}}(u, \tilde{\tau}) \rangle & \dots  \\
\langle \partial_2 \mathcal{E}_{\theta_{\mathcal{E}}}(u, \tilde{\tau}), \partial_1\mathcal{E}_{\theta_{\mathcal{E}}}(u, \tilde{\tau}) \rangle & || \partial_2 \mathcal{E}_{\theta_{\mathcal{E}}}(u, \tilde{\tau}) ||_2^2 & \dots \\ 
\vdots & \vdots & \ddots
\end{pmatrix} .
\end{align}

\vspace{-4mm}
\begin{wraptable}{R}{0.52\textwidth}
\vspace{-4.0mm}
\setcounter{table}{1}
\begin{tabular}{wl{2.2cm} | P{2.5cm} | P{2.5cm} } 
\toprule
\belowrulesepcolor{light-gray} 
\rowcolor{light-gray} \multicolumn{3}{l}  {1-d Burger's noise-added results} 
\\ 
\aboverulesepcolor{light-gray} 
\midrule
{method}  & {$\mathcal{C}_{\text{Burger's}}^{1}, t=0.35$} & {$\mathcal{C}_{\text{Burger's}}^{2}, t=0.35$}
\\ 
\midrule
\rowcolor{LightCerulean} {Ricci flow}  & {$25.9 \pm 20.0$} & {$12.4 \pm 7.70$} 
\\ 
{GD-VAE}  & {$49.7 \pm 131$}  & {$34.5 \pm 40.6$}   
\\
{AE}  & {$14.8 \pm 6.64$}  & {$15.6 \pm 10.4$}  
\\
{AE-overtrained}  & {$19.4 \pm 9.74$}  & {$20.8 \pm 12.4$}
\\
{AE-extended}  & {$20.1 \pm 10.2$}  & {$22.6 \pm 20.2$}  
\\
\bottomrule

\end{tabular}
\vspace{2mm}

\begin{tabular}{wl{2.2cm} | P{2.5cm} | P{2.5cm} } 
\toprule
\belowrulesepcolor{light-gray} 
\rowcolor{light-gray} \multicolumn{3}{l}  {1-d diffusion-reaction noise-added results} 
\\ 
\aboverulesepcolor{light-gray} 
\midrule
{method}  & {$\mathcal{C}_{\text{diffusion}}^{1}, t=0.35$} & {$\mathcal{C}_{\text{diffusion}}^{2}, t=0.35$}
\\ 
\midrule
\rowcolor{LightCerulean} {Ricci flow}  & {$18.6 \pm 11.0$} & {$17.1 \pm 9.48$} 
\\ 
{GD-VAE}  & {$13.2 \pm 7.68$}  & {$13.6 \pm 11.6$}   
\\
{AE}  & {$63.6 \pm 30.8$}  & {$59.8 \pm 37.9$}  
\\
{AE-overtrained}  & {$51.3 \pm 31.7$}  & {$39.0 \pm 32.8$}  
\\
{AE-extended}  & {$44.5 \pm 21.5$}  & {$81.6 \pm 81.4$}  
\\
\bottomrule

\end{tabular}
\vspace{2mm}
\caption{We examine relative $L^1$ error with initial data corrupted with noise on 30 test examples. Observe the Ricci flow method is fairly consistent among the methods. The high variances in the $\mathcal{C}_{\text{Burger's}}^1$ and $\mathcal{C}_{\text{diffusion}}^2$ settings are partially due to outliers (recall in the Burger's experiment, $\alpha, \beta \in [-1,1]$, which allows near zero cases).}
\label{tab:noise_injected_data}

\vspace{-0.5mm}

\vspace{5mm}
\begin{centering}
\begin{tabular}{wl{2.2cm} | P{4.8cm} } 
\toprule
\belowrulesepcolor{light-gray} 
\rowcolor{light-gray} \multicolumn{2}{l}  {1-d Burger's extrapolation} 
\\ 
\aboverulesepcolor{light-gray} 
\midrule
{method}  & {$\mathcal{B}_{\text{Burger's}}^{1}, t=0.35$}
\\ 
\midrule
\rowcolor{LightCerulean} {Ricci flow}  & {$8.79 \pm 4.84$} 
\\ 
{GD-VAE}  & {$36.1 \pm 17.8$}   
\\
{AE}  & {$24.7 \pm 20.3$}  
\\
{AE-overtrained}  & {$21.2 \pm 12.7$}  
\\
{AE-extended}  & {$40.4 \pm 18.6$}  
\\
\bottomrule

\end{tabular}
\vspace{0mm}
\caption{We examine relative $L^1$ error for out-of-distribution data. We list $\mathcal{B}_{\text{Burger's}}^1$ in Appendix \ref{appendix_extrap_sets}.}
\label{burgers_extrap_errors}
\end{centering}

\vspace{-8mm}
\end{wraptable}

The expectations run over time and the training data, taking data $\phi_0^i \sim \Phi$ for PDE data belonging to distribution $\Phi$. We use notation $|\mathcal{U}| = \dim(\mathcal{U})$. To clarify, $\tilde{\phi}_t = \phi(\cdot, t)$ is the discretization of $\phi$ at time $t$. Let $\Theta = (\Theta_{u}, \Theta_g, \Theta_{\mathcal{E}}, \Theta_{\mathcal{D}})$.  Let $\hat{\tau} \sim U(\{ C_T i \Delta t : i \in \mathbb{N}, C_T, \Delta t \in \mathbb{R}^+ \} )$ be a uniform sampling on a scaling by constant $C_T$ of the time domain $[0,T]$ discretized (generally $C_T=1$), and let $\tilde{\tau} \sim U([0, \tau'])$ be uniformly sampled via a collocation procedure. We generally choose $\hat{\tau}, \tilde{\tau}$ to correspond by an identity relationship. Note that all times $t, \hat{\tau}, \tilde{\tau}$ are associated by some constant scaling. Hence, a time in the ambient PDE evolution has a corresponding time along the Ricci flow evolution to be decoded. The possible scaling of times helps facilitate a robust manifold representation, i.e. far in its evolution from a singularity. $||\cdot||_F$ is the Frobenius norm. All derivatives are computed via automatic differentiation \citep{baydin2018automatic}. The analytic expectations can be taken using empirical averages, additionally divided into batches with parameters updated with a stochastic optimizer \citep{kingma2017adam}.

\vspace{2mm}

The loss terms are organized as a Ricci flow physics-informed term $\mathcal{L}_{Ric}$, quadratic loss $\mathcal{L}_{dec}$ to ensure the decoder matches the PDE solution, and a constraint term $\mathcal{L}_{met}$ to match the manifold evolution with the metric satisfying Ricci flow. $\lambda_{dec}$ and $\lambda_{met}$ are scaling coefficients as needed (generally $\lambda_{dec} = \lambda_{met} = 1$). In the final term $\mathcal{L}_{met}$, $g_{\theta_g}$ is the matrix of the physics-informed metric term $g_{\theta_g}$, where the components are summed with respect to each inner product $\langle \partial_j \mathcal{E}_{\theta_{\mathcal{E}}}, \partial_k \mathcal{E}_{\theta_{\mathcal{E}}} \rangle $ to enforce loss for each entry.

\vspace{2mm}

\textbf{Special cases.} Special cases of Ricci flow may be under consideration, which are favorable for reasons such as lower computational cost or higher-dimensional encodings. In these scenarios, information about the manifold is set beforehand, such as the metric. Learned latent geometries are sacrificed at the gain of some other benefit. We discuss special cases in Appendix~\ref{special_cases}. Coordinate transformations may be appropriate under select special cases. Our method in such a setting is discussed in Appendix~\ref{coord_transform}.

\section{Experiments}

We present Ricci flow-guided autoencoders on a series of numerical experiments for approximate solutions of PDEs given collections of known data. We compare our method to existing GD-variational autoencoder methods in \citep{lopez2022gdvaes}, as well as other forms of autoencoder baselines. Results for both the Ricci flow method and alternative methods are collected using identical initial condition input spaces. The GD-VAE method is performed using an evolution mapping $\phi_{t} \rightarrow \phi_{t + \tilde{t}}$, where $\tilde{t}$ is some time-increment. Predictions are retrained for new values of $\tilde{t}$. We emphasize our method needs no retraining for new $t$; the latent manifold holds for all times in the ambient PDE.

\begin{table*}[ht]
\small
\centering
\begin{tabular}{wl{2.0cm} |  P{2.1cm} P{2.1cm} P{2.1cm} P{2.1cm} P{2.1cm}} 
\toprule
\belowrulesepcolor{light-gray} 
\rowcolor{light-gray} \multicolumn{6}{l}  {1-d Burger's equation} 
\\ 
\aboverulesepcolor{light-gray} 
\midrule
{method} & {$t=0$} & {$t=0.25$} & {$t=0.5$} & {$t=0.75$} & {$t=1$} 
\\ 
\midrule
\rowcolor{LightCerulean}
{Ricci flow}  & {$2.22 \pm 1.31$} & {$1.51 \pm 0.647$} & {$1.69 \pm 0.754$} & {$1.50 \pm 0.778$} & {$2.01 \pm 1.04$}
\\ 
{GD-VAE}  & -   & {$14.4 \pm 18.9$}  & {$4.52 \pm 5.47$}   & {$4.90 \pm 6.88$} & {$3.75 \pm 3.68$}   
\\
{AE, g-proj.}  & -   & {$10.3 \pm 10.6$} &  {$10.4 \pm 7.89$}   & {$5.29 \pm 11.9$}  & {$2.24 \pm 1.13$}  
\\

\bottomrule

\end{tabular}

\vspace{2mm}

\begin{tabular}{wl{2.0cm} |  P{2.1cm} P{2.1cm} P{2.1cm} P{2.1cm} P{2.1cm}} 
\toprule
\belowrulesepcolor{light-gray} 
\rowcolor{light-gray} \multicolumn{6}{l}  {1-d diffusion-reaction}  
\\ 
\aboverulesepcolor{light-gray} 
\midrule
{} & {$t=0$} & {$t=0.25$} & {$t=0.5$} & {$t=0.75$} & {$t=0.9$}
\\ 
\midrule
\rowcolor{LightCerulean} 
 {Ricci flow}  & {$0.274 \pm 0.136$} & {$0.184 \pm 0.0755$} & {$0.215 \pm 0.0932$} & {$0.429 \pm 0.199$}  & {$1.14 \pm 0.410$}
\\
{GD-VAE}  & -   & {$2.10 \pm 2.81$} & {$2.94 \pm 6.17$}   & {$1.53 \pm 0.886$}  & {$2.24 \pm 0.881$}
\\
{AE, g-proj.}  &  -   & {$1.78 \pm 2.59$} &  {$1.28 \pm 0.903$}   & {$1.41 \pm 0.595$}  & {$2.77 \pm 1.33$}
\\
\bottomrule

\end{tabular}

\vspace{2mm}

\hspace{0.15mm}
\begin{tabular}{wl{2.5cm} |    P{3.6cm}   P{3.6cm}   P{3.6cm}} 
\toprule
\belowrulesepcolor{light-gray} 
\rowcolor{light-gray} \multicolumn{4}{l}  {1-d diffusion-reaction extrapolation}  
\\ 
\aboverulesepcolor{light-gray} 
\midrule
{}  & {$\mathcal{B}_{\text{diffusion}}^{1}, t=0.35$} & {$\mathcal{B}_{\text{diffusion}}^{2}, t=0.35$} & {$\mathcal{B}_{\text{diffusion}}^{3}, t=0.35$}
\\ 
\midrule
\rowcolor{LightCerulean} 
 {Ricci flow}  & {$10.4 \pm 15.4$}  & {$ 17.8 \pm 9.62$}   & {$30.8 \pm 13.1$}
\\
{GD-VAE}  & {$14.9 \pm 17.0$}   & {$21.9 \pm 13.9$}   & {$36.1 \pm 18.6$}
\\
{AE}  & {$19.4 \pm 27.0$}   & {$23.6 \pm 12.8$}   & {$47.5 \pm 26.3$}
\\
{AE-overtrained}  & {$7.85 \pm 12.2$}   & {$22.3 \pm 10.1$}   & {$40.8 \pm 16.1$}
\\
{AE-extended}  & {$20.8 \pm 31.6$}   & {$21.7 \pm 11.1$}   & {$39.2 \pm 15.6$}
\\
\bottomrule

\end{tabular}
\label{diffusion_reac_extrap}

\vspace{2mm}

\begin{tabular}{wl{2.0cm} | P{2.1cm} P{2.1cm} P{2.1cm} P{2.1cm} P{2.1cm}}
\toprule
\belowrulesepcolor{light-gray} 
\rowcolor{light-gray} \multicolumn{6}{l} {2-d Navier-stokes equation}  
\\ 
\aboverulesepcolor{light-gray} 
\midrule
{} & {$t=0$} & {$t=0.25$} & {$t=0.5$} & {$t=0.75$}   & {$t=1.0$} 
\\ 
\midrule
\rowcolor{LightCerulean} 
 {Ricci flow}  & {$4.44 \pm 1.48$}  & {$3.97 \pm 1.57$}   & {$5.26 \pm 1.18$} & {$8.46 \pm 1.09$} & {$9.25 \pm 1.53$} 
\\
\bottomrule

\end{tabular}

\caption{We compare our methodology to those existing on in-distribution and out-of-distribution scenarios, reporting relative $L^1$ errors on 30 testing sets. All errors are taken $10^{-2}$. Times corresponding to the increment $\tilde{t} = 0$ are omitted for comparisons since this is a trivial learning task in these frameworks. Such error by a decrease in order of magnitude for the diffusion-reaction experiment has partial attribution to the enhanced MLP architecture (Appendix~\ref{MMLP}).}

\end{table*}

\subsection{Viscous Burger's equation}


We first introduce our method on the viscous Burger's equation. The viscous Burger's equation is given by
\begin{equation} 
\partial_t \phi(x,t) + \phi(x,t) \partial_x \phi(x,t) = \nu \partial_x^2 \phi(x,t) ,
\\
{}
\end{equation}
for diffusion coefficient parameter $\nu \in \mathbb{R}$ \citep{lopez2022gdvaes}. Our numerical evaluations for training and test data were constructed using a psueudo-spectral method with Fast Fourier transforms \citep{burgerseqgithub}. Evaluations were over a $(x,t) \in [0,1]\times[0,1]$ domain discretized into an equispaced grid $\Omega_x \times \Omega_t$ with mesh distance $h_x=h_t=0.01$. Kinematic viscosity coefficient $\nu=0.01$ was chosen. Initial conditions were evaluated from $\mathcal{A}_1 = \{ \phi_0 : \phi_0 = \phi(x,0) = \alpha \sin(2 \pi x) + \beta \cos^3(2 \pi x), \alpha, \beta \in [-1,1] \} .$ We choose training data by uniformly sampling 100 of $\phi_t$ for $t \in \Omega_t$ per each $\phi_0$. We set $\mathcal{U} \subseteq \mathbb{R}^m = \mathbb{R}^2$, and such a low dimension reduces offline computational cost.

\begin{figure*}
  \vspace{0mm}
  \centering
  \includegraphics[scale=0.65]{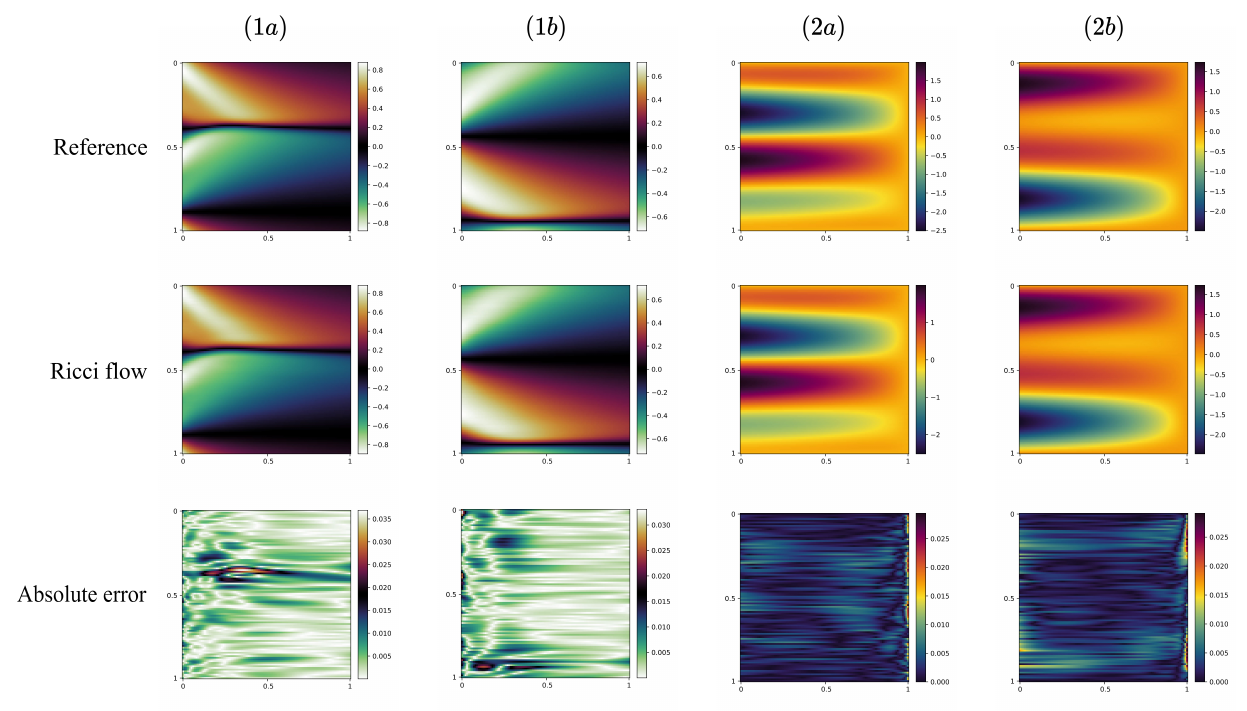}
  \caption{Illustration of numerical $(1)$ Burger's equation and $(2)$ diffusion-reaction solutions versus their Ricci flow solutions on in-distribution scenarios. $(a)$ and $(b)$ are two distinct solutions.}
  \label{fig:burgers_eq_ricci_flow}

  \vspace{10mm}
  \centering
  \includegraphics[scale=0.65]{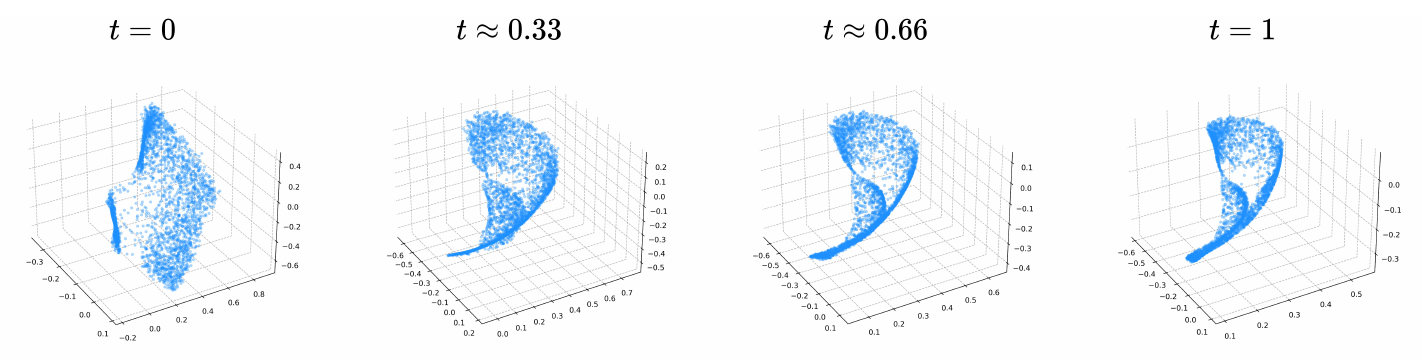}
  \caption{Illustration of the manifold latent space subject to Ricci flow in the Burger's equation experiment. The unusual behavior at $t=0$ is potentially because it is a boundary, and is more difficult to learn. We provide an alternative view in Appendix \ref{app:burgers_eq_manifold_app}.}
  \label{fig:burgers_eq_ricci_flow}
\end{figure*}

\begin{figure}[htbp]
  \vspace{0mm}
  \centering
  \includegraphics[scale=0.44]{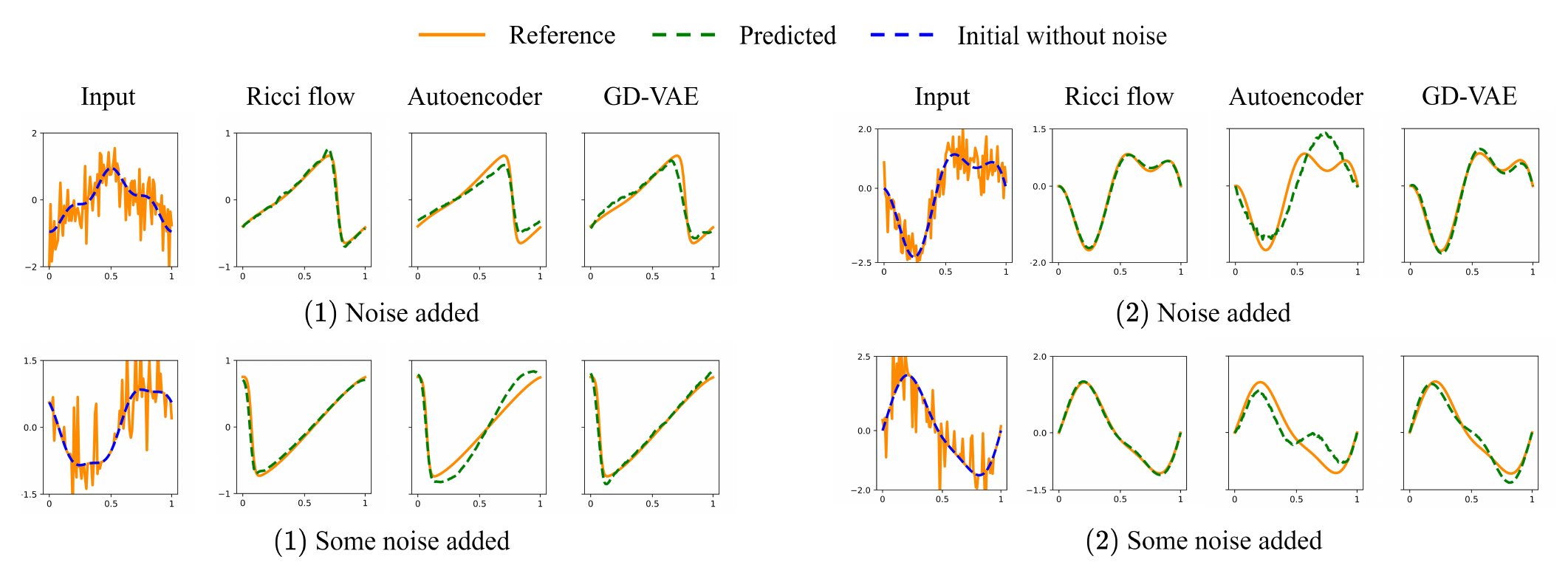}
  \caption{We compare methodology in the setting with injections of noise in the $(1)$ Burger's equation and $(2)$ diffusion-reaction experiments. We examine $t=0.35$.}
  \label{fig:noise_burgers_and_diffreac}
\end{figure}

\vspace{2mm}

\vspace{2mm}

Training consisted of a generation of $3,000$ initial conditions. Loss coefficients $\lambda_{dec} = \lambda_{met} = 1$ were taken. We found training to occasionally be unstable, individuated by sudden increases in loss by orders of magnitude, resolvable by using a low learning rate, weight decay, and gradient clipping. The weight decay parameter was reduced in late training to ensure objective satisfaction.

\vspace{2mm}

Width of $w=120$ was chosen for all neural networks in this experiment. All neural networks held 3 hidden layers. Experimentation for activation was upon both $\text{GELU}(\cdot)$ \citep{hendrycks2023gaussian} and $\text{tanh}(\cdot)$, a choice of which was nonvital; however, we remark activation with sufficient continuous differentiability properties are necessary, as second order derivatives are taken to formulate the physics-informed Ricci flow. $C^{\infty}$ activations also ensure smoothness when forming the manifold, which helps support our embedding dimension choice (see section \ref{sec:embedding_dims}).

\vspace{2mm}

We found success in taking $\text{sigmoid}(\cdot)$ activation in the final layer of network $u_{\theta_u}$, scaled by some arbitrary constant (we choose $2\pi$). This forces the learned parameterization domain to be relatively small in measure, which allows highly nonzero partial derivatives to formulate the Riemannian metric from the manifold. This helps train both the manifold and the physics-informed network.

\vspace{2mm}

\begin{wrapfigure}{R}{0.5\textwidth}
  \centering
  \includegraphics[width=0.95\linewidth]{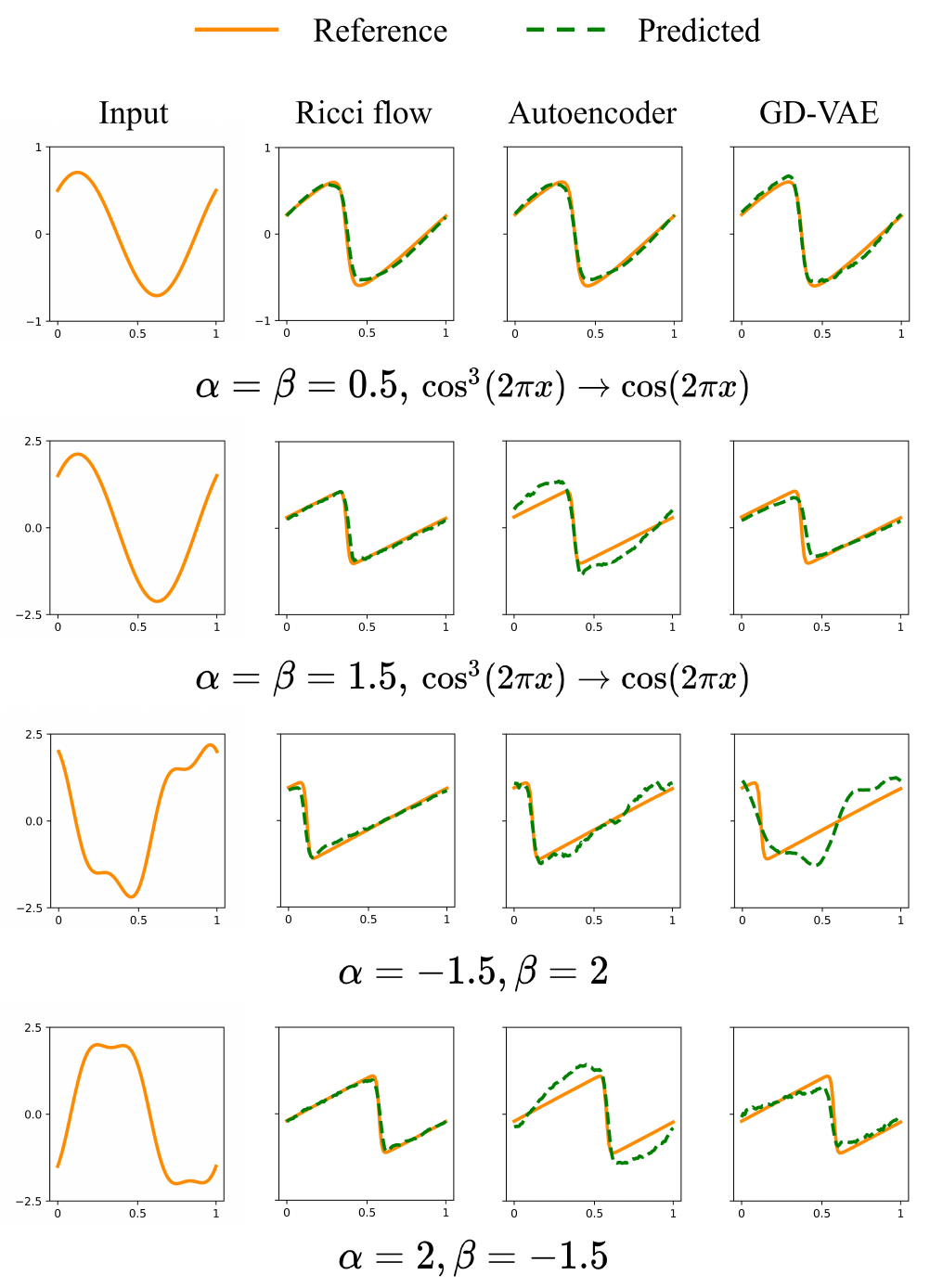}
  \vspace{-0.65mm}
  \caption{We compare various methodology in the extrapolation setting of out-of-distribution input data of varying difficulty for Burger's equation. We examine $t=0.35$.}
  \label{fig:burgers_extrapolation}
  \vspace{-5.5mm}
\end{wrapfigure}

We examine our methodology on out-of-distribution cases belonging to the same training family, but with parameter deviations. We denote these sets as $\mathcal{B}$. In Figure~\ref{fig:burgers_extrapolation}, we give a comparison of our methodology, listing the changes made to the data. We list extrapolation error results in Table~\ref{burgers_extrap_errors}. Our extrapolation baselines are the GD-VAE framework, and a vanilla autoencoder with hyperparameters matching that of the Ricci flow setting when applicable, including a latent space of $\mathbb{R}^3$. The AE g-proj. autoencoder as presented in \citep{lopez2022gdvaes} is the autoencoder baseline in the in-distribution test setting, maintaining consistency between manifold-based methods. 

\vspace{2mm}

The AE baseline is trained with comparable in-distribution training error to what we found in the Ricci flow case. The AE-overtrained error is an autoencoder baseline where training is continued convergence, which helps invoke the grokking phenomenon. The AE-extended architecture is an autoencoder with exact architecture as our Ricci flow method, but without the latent manifold. This means we map to a $\mathcal{U}$ once and subsequently low-dimensional Euclidean space (typically $\mathbb{R}^3$) before it is decoded. This comparison is meaningful because it removes error differences that are possibly from architecture, and we investigate performance differences due to the manifold and Ricci flow exactly. We remark we map to $\mathcal{U} \subseteq \mathbb{R}^3$ instead, which trains better than over $\mathbb{R}^2$, and we include $\text{sigmoid}(\cdot)$ in the last layer for $\mathcal{U}$ in this experiment only. Our baselines were trained with near identical activation and learning rate when applicable.

\vspace{2mm}

In Figure ~\ref{fig:noise_burgers_and_diffreac}, we introduce noise to the data in a robustness experiment. We introduce Gaussian noise among all points in the discretization with variance $\sigma^2 = 0.5$, the set of which we denote $\mathcal{C}^1$, and in sparse locations with $\sigma^2=0.75$, which we denote $\mathcal{C}^2$. We found large learning rate ($5\text{e}{-4}$) had performance improvement in robustness in the GD-VAE setting in this experiment. We illustrate empirical error results corresponding to the same two scenarios in table \ref{tab:noise_injected_data}. Refer to table \ref{tab:noise_injected_data} for comprehensive results on numerous instances of data, as figure \ref{fig:noise_burgers_and_diffreac} is too limited to present a wide overview of error analysis.

\subsection{Diffusion-reaction equation}

\vspace{-0mm}

\afterpage{

\begin{figure*}[t]
  \vspace{0mm}
  \centering
  \includegraphics[scale=0.72]{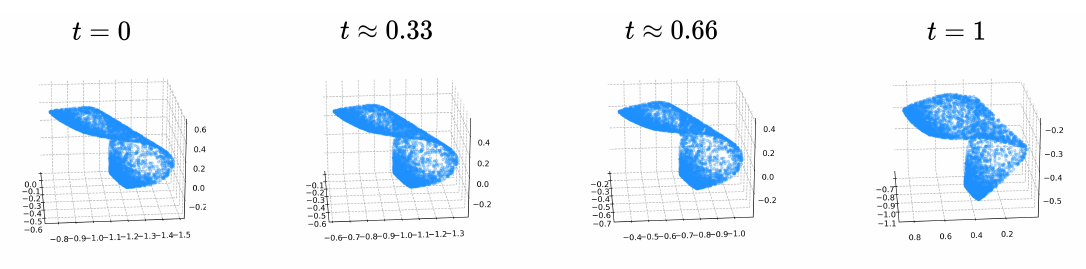}
  \caption{Illustration of the manifold latent space subject to Ricci flow in learning diffusion-reaction PDE data.}
  \label{fig:training_loss_late_training}
\end{figure*}
\vspace{5mm}
}

In this experiment, we study the setting of the 1-dimensional diffusion-reaction PDE with source term, given by the equation \citep{wang2021learning}
\begin{equation}
\partial_t \phi(x,t) = D  \partial_x^2 \phi(x,t) + \lambda \phi^2(x,t) + f(x) ,
\end{equation}
for diffusion coefficient $D \in \mathbb{R}$, reaction rate $\lambda \in \mathbb{R}$, and source term $f$. Our numerical scheme for constructing data was a finite difference solver \citep{wang2021learning}. Datasets were constructed over a $(x,t) \in [0,1]\times[0,1]$ domain. We choose mesh distance $h_x = h_t = 0.01$. We generated initial data from $f(x)$, using a special case of a Fourier series, $\mathcal{A}_2 = \{ f : f(x) = \alpha \sin(2 \pi x) + \frac{\alpha+0.5}{2} \cos( 4 \pi x) + \frac{\beta}{3} \sin( 4 \pi x), \alpha, \beta \in [-1,1] \}$. The sum taken in the second coefficient ensures non-triviality of the initial data. Training data is normalized so that $\int_{\mathcal{X}} |\phi_0| dx = 1$, and the same scaling coefficient also scales data at time $t>0$. We learn the mapping $(\phi_0,t) \rightarrow \phi_t$ despite initial functions being sampled for $f$ and not $\phi_0$. Again, we set $\mathcal{U} \subseteq \mathbb{R}^m = \mathbb{R}^2$.

\vspace{2mm}

We implement an augmented architecture, the modified multilayer perceptron (MMLP), in this experiment, which has demonstrated success in physics-informed settings \citep{wang2021learning}. This architecture is elaborated upon in Appendix~\ref{MMLP}, with training loss illustrated in Appendix~\ref{fig:loss_comparison_fig}, indicating this architecture achieves an order of magnitude lower training loss in fewer iterations of the Adam optimizer. In Table~\ref{diffusion_reac_extrap}, our autoencoder baseline in the extrapolation setting uses the architecture used in the Ricci flow setting to allow fair comparison despite the baseline autoencoder being a non-physics-informed setting.

\vspace{2mm}

Training consisted of $5,000$ samples from $A_2$ paired with $100$ uniformly sampled evaluations at later times $\phi_0(\cdot, t^*)$. Architecture consisted of $5$ hidden layers for each network, a hidden layer being the term $\zeta^i$ as illustrated in Appendix~\ref{MMLP}, which had a width of $100$. $\text{GELU}(\cdot)$ was used for all activations. We do not take $\text{sigmoid}(\cdot)$ activation in the final layer of $u_{\theta_u}$. We found the learned domain naturally has relatively small measure (see Appendix \ref{fig:param_domain_diffusion}), hence sufficiently large derivatives. 

\vspace{2mm}

As with Burger's equation, we provide an experiment with noise introduced into the data in various settings. We illustrate results in figure ~\ref{fig:noise_burgers_and_diffreac} and table ~\ref{tab:noise_injected_data}.

\setcounter{figure}{5}
\begin{figure}[htbp]
  \vspace{0mm}
  \centering
  \includegraphics[scale=0.66]
  {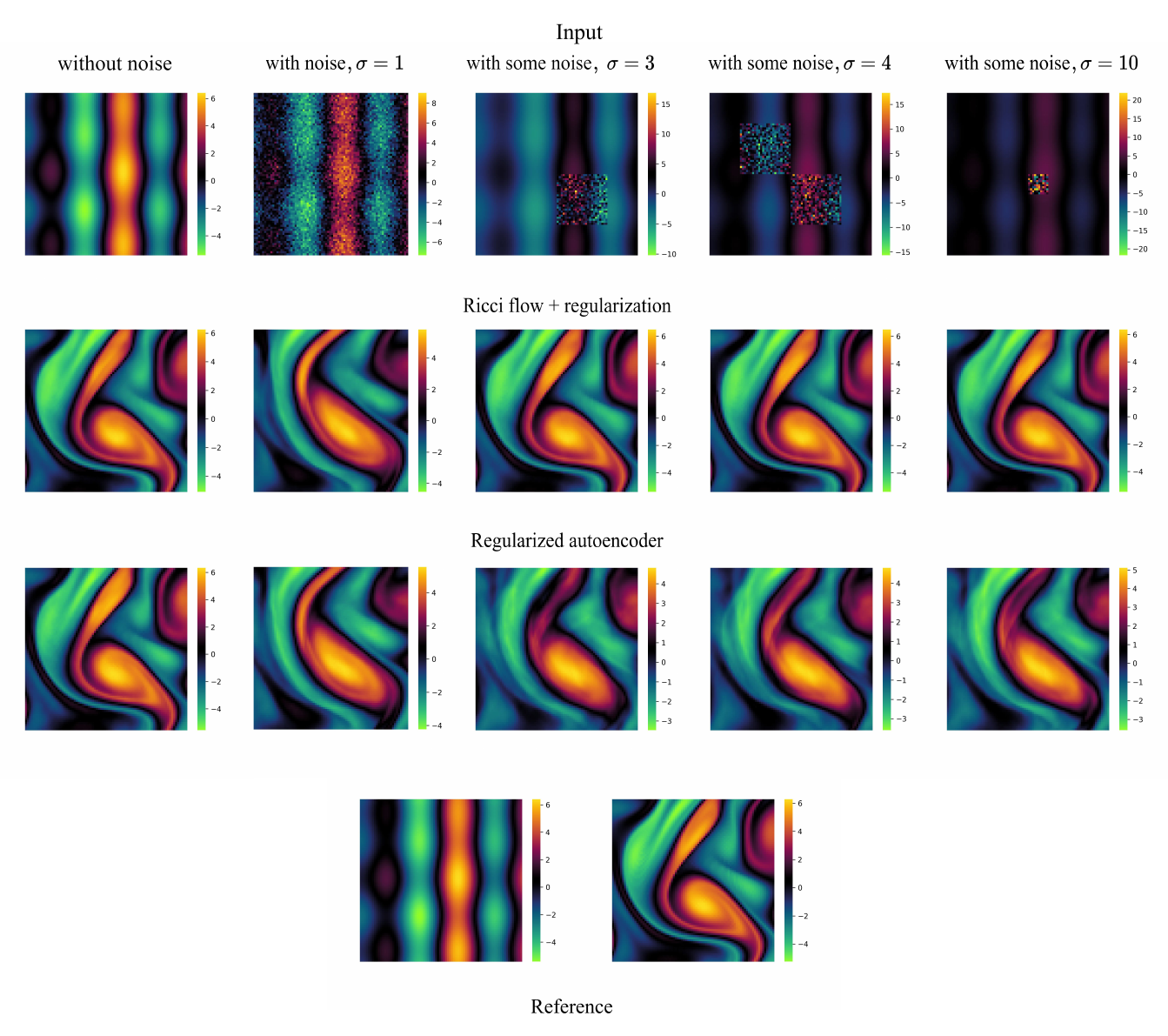}
  \vspace{2mm}
  \caption{We illustrate results on the Navier-Stokes experiment with noise added using an autoencoder baseline. We set the seed for identical noise. We hold latent dimension $d=100$ for the autoencoder, use similar architectures as to our method, and use dropout. In the Ricci flow case, we enforce Ricci flow on the sphere via $\text{radius}(\tau) \cdot u/||u||$, whereas the regular autoencoder case has no manifold enforcement.}
  \label{fig:navier_stokes_noise}
\end{figure}

\subsection{Navier-Stokes equation}
\label{Navier_Stokes_section}

We test our method on the 2-d incompressible Navier-Stokes equation \citep{crewsincompressibleNS},
\begin{equation}
\partial_t v(x,t) + (v(x,t) \cdot \nabla) v(x,t) - \nu \nabla^2 v(x,t) = \frac{-1}{\rho} \nabla p(x,t) ,
\end{equation}
given by the velocity field $v$ evolution subject to the constraint $\nabla \cdot v = 0$, which is the incompressibility condition, and  given an initial velocity vector field $v_0$ and pressure field over grid $\Omega_x \subseteq \mathcal{X} \subset \mathbb{R}^2$. The vorticity $\phi = \nabla \times v$ serves as training data, where $\nabla \times$ denotes the curl. We learn the autoencoder mapping $[(\nabla \times v_0)]_z \times t \rightarrow \phi_t$, where $[\cdot]_z$ is the $z$-th coordinate over $\Omega_x$. Data for this experiment was constructed using a discontinuous Galerkin/Fourier spectral solver \citep{crewsincompressibleNS}
\citep{crewsnavierstokes}.

\vspace{2mm}

\begin{wrapfigure}{R}{0.5\textwidth}
\vspace{0.0mm}
  \vspace{-2.5mm}
  \centering
  \includegraphics[scale=0.44]{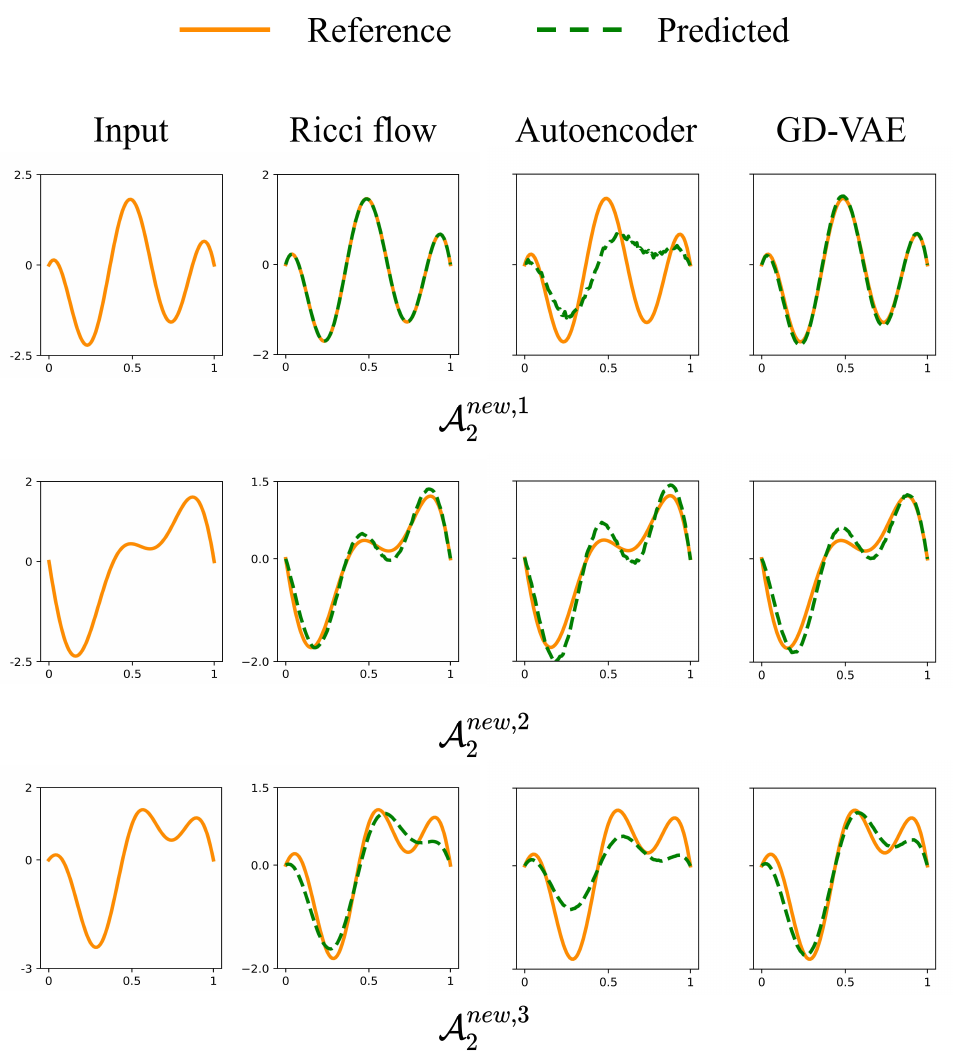}
  \vspace{-0.35mm}
  \caption{We compare various methodology in the extrapolation setting for the diffusion-reaction equation. We examine $t=0.35$.}
  \label{fig:training_loss_late_training}
  \vspace{-0mm}
\end{wrapfigure}

We consider the special case of Ricci flow upon the $(d-1)$-sphere. Data for this experiment consisted of $80 \times 80$ meshes in $\mathbb{R}^2$, data of markedly higher dimensionality than in the previous experiments. A parameterization representation in $\mathbb{R}^2$ and an embedding in $\mathbb{R}^3$ were empirically found too limiting through dimensionality to thoroughly represent the data, and information was lost through such compression to accurately reconstruct solutions as well as provide prediction capacity. An encoding onto the $(d-1)$-sphere allows for a dimension increase in $\mathcal{U}$ and the embedding space $\mathbb{R}^d$ by a reduction in the numerical power needed. While the manifold still evolves under a closed form of Ricci flow, the geometries are now fixed, but the latent expressivity is of enough importance to make the evolving $(d-1)$-sphere more suitable.

\vspace{2mm}

A convolutional neural network (CNN) \citep{oshea2015introduction} was used for the parameterization network $u_{\theta_{u}}$, taking 2-d grid input and mapping to $\mathcal{U} \subseteq \mathbb{R}^{100}$ for manifold parameterization. The $(d-1)$-sphere was embedded in $\mathbb{R}^{101}$. We observed an artificial neural network gained greater empirical success over a transpose layer-based CNN for the decoder, achieving lower training error in fewer iterations of the Adam optimizer. Architecture for the decoder was taken as $101-400-1600-6400-6400-\text{(output)}$, an architecture susceptible to overfitting. We applied techniques to prevent such, as described as follows. Wide architecture with severe overfitting regularization gained empirical favor over narrow architecture with modest regularization in the test setting.

\vspace{2mm}

Overfitting techniques were conducive towards generalization and accurately reconstructing solutions in the test setting. These techniques were notably dropout and adding Gaussian noise to the intermediate steps in producing the output, such as the parameterization space and the manifold in the embedded space \citep{noiselatentspace_stack}. Dropout was only used in the decoder, set in earlier layers as $p=0.325$ and reduced before the final layer to $p=0.225$. For the Gaussian noise, we first took a local coordinate $u \in \mathcal{U}$ and recast it as $u + C_{\xi^u} \cdot \xi^u $, where we use $[\xi^u]_{i} \sim \mathcal{N}(0,1)$ to denote a tensor with standard normal elements $[\cdot]_{i}$ with the same dimensions as $u$. $C_{\xi^u}$ denotes a scaling constant. We also considered adding Gaussian noise to manifold of the form $p_{\text{new}} = p_{\mathcal{M}} + C_{\xi^{\mathcal{M}}} \cdot r(\tau) \cdot \xi^{\mathcal{M}} $, instead scaling with the radius here and denoting $\xi^{\mathcal{M}}$ the analog of $\xi^u$ for the manifold. Overall, we found both successful, but stuck with adding noise to local coordinates.

\vspace{2mm}

The introduction of latent noise in this experiment acts similarly to a variational framework by transforming the latent space to a probabilistic setting, but training is not done with an ELBO-type loss, and the framework is missing desirable qualities that could otherwise be present at the gain of generalization. For example, the addition of noise comes at a sacrifice of structure, as the training data now lies in regions locally about but not on the manifold; however, the test data is evaluated without the addition of noise, and so test data is exactly along a manifold. Also, there is a loss of identifiability: This is observable in figure \ref{fig:navier_stokes_noise}, as severely-noisy ambient PDE data can be inferred as entirely different solutions, and so similar regions in latent space cannot entirely be inferred as closely related in the ambient space.

\vspace{2mm}

Training data generation was costly for this experiment, taking about $\sim8$ hours to generate 2200 solutions, 2000 for training and 200 for test. Such an increase in data is a less viable option to reduce overfitting, as it is often a limited option in practice. 

\vspace{2mm}

The GD-VAE framework was established for this experiment using the  architecture as presented in the repository, but width and depth were expanded to accommodate the increased difficulty of learning the data. Deep artificial neural networks were used for both the encoder and decoder. In practice, this methodology without heavy modification presented itself as unsuitable for this experiment, likely due to latent dimensionality.

\section{Additional discussion}

We discuss possibilities as to why our methods help robustness. Since void topologies are not unusual under unregularized losses, a partial consequence of this is that out-of-distribution data has lower likelihood to deviate from its true latent point with our regularized methods. For example, in Figures \ref{fig:param_domain_noreg}, \ref{fig:param_domain_noreg_2}, out-of-distribution data strays further than its in-distribution data compared to its analogous figure of Figure \ref{fig:param_domain_burgers}. We also observe near singularities and twisted geometries in \ref{fig:noreg_manifold}, \ref{fig:noreg_manifold_2}. If the metric collapses, fixed local distances actually grow in inverse proportion to a measure of the geometry. For example, a fixed distance locally appears smaller along a large manifold compared to a small one. Thus, metric collapse can distort how out-of-distribution data behaves. Since Ricci flow is a curvature flow, its curvature can expand in norm as the metric approaches singularity, helping ensure nondegenerate metrics. A canonical geometry can also help ensure the manifold is sufficiently "round." For example, it is possible a manifold forms with measure with respect to one direction of the frame much smaller than another direction, i.e. a "long" manifold or a "squeezing" effect. Thus, an out-of-distribution point can more easily fall off the manifold. Another potential reason is that curvature uniform metrics preserve geodesics more consistently, helping predictability. Our code is available at \url{https://github.com/agracyk2/Ricci-flow-guided-autoencoders-for-dynamics}.

\section{Future work and limitations}

Additional work could be created that seeks a method to incorporate extrapolation and noise data into the manifold in a more effective way. An implicitly-learned manifold undergoing dynamics would also be of consideration, in which such an accommodation could in turn improve such results even further. It can be noted \citep{lopez2022gdvaes} suggests certain qualities such as periodicity work together with certain latent geometries. The concept of optimality of geometry could be further analyzed. One may attempt to quantify the question: how do deep neural networks harmonize well with data in a sense of geometry, particularly a geometric latent space? Physics-informed Ricci flow is a nontrivial training task from a standpoint of computational difficulty. It is of interest to refine such a training algorithm to make this expedited and more practicable in an offline setting.

\section{Acknowledgments}

I would like to thank Xiaohui Chen for helpful discussions in developing this project, particularly regarding intrinsic versus extrinsic geometric flows. Andrew Gracyk was supported by NSF under Grant No. 1922758 from DIGIMAT sponsorship.

\bibliography{ricci_flow_autoenc}

\appendix

\newpage

\section{Navier-Stokes figure}

\setcounter{figure}{5}
\begin{figure}[htbp]
  \vspace{0mm}
  \centering
  \includegraphics[scale=0.66]
  {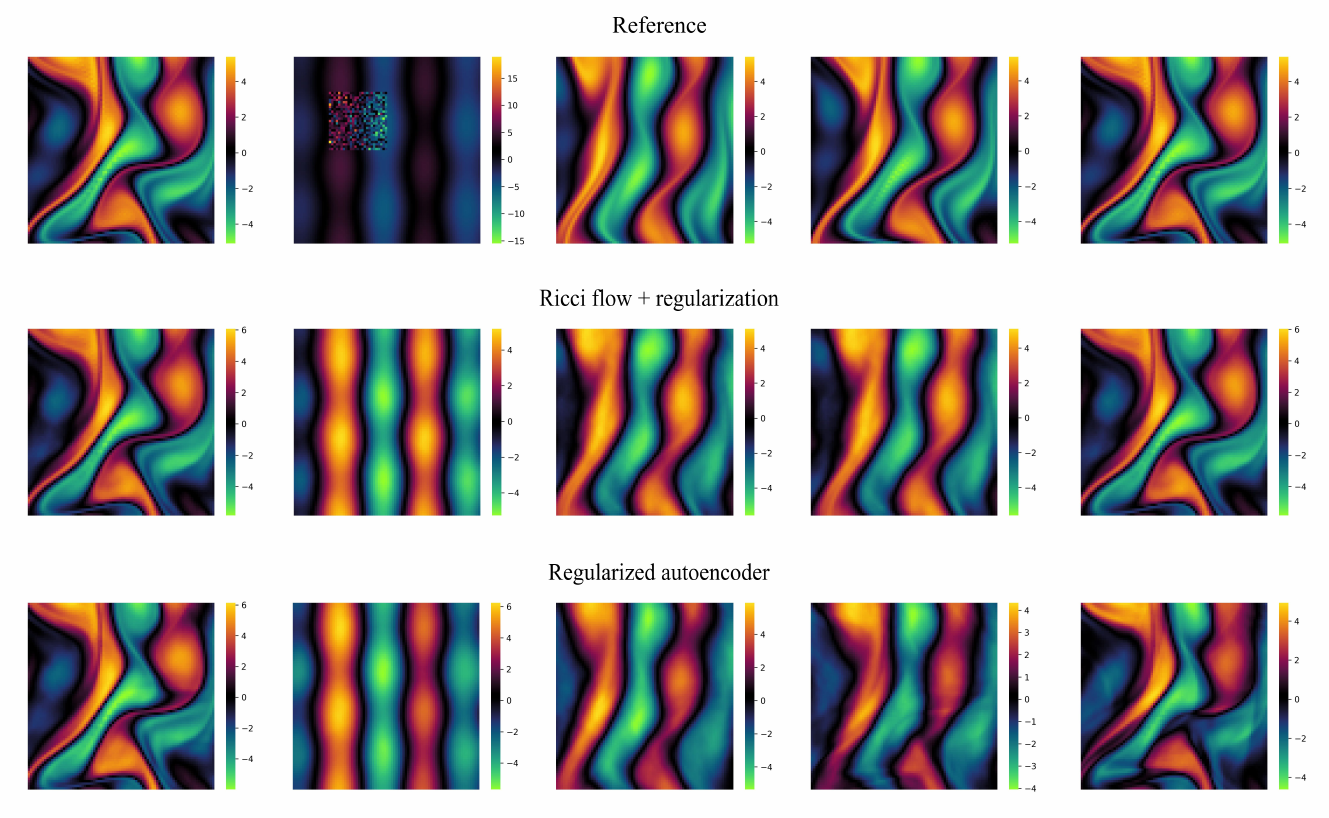}
  \vspace{2mm}
  \caption{We illustrate results on the Navier-Stokes experiment with noise $\sigma=4.0$ added using an autoencoder baseline. We set the seed for identical noise. We hold latent dimension $d=100$ for the autoencoder, use similar architectures as to our method, and use dropout. In the Ricci flow case, we enforce Ricci flow on the sphere via $\text{radius}(\tau) \cdot u/||u||$, whereas the regular autoencoder case has no manifold enforcement. In this experiment, we plot 3 time increments late in the learning. It can be noted the last time illustrated has greater proportionality of containment in the training set. The first column is corresponds to the final time increment without noise, and the last with noise.}
  \label{fig:navier_stokes_noise_progression}
\end{figure}

\section{2-d wave equation experiment}

\begin{figure*}
  \centering
  \includegraphics[scale=0.65]{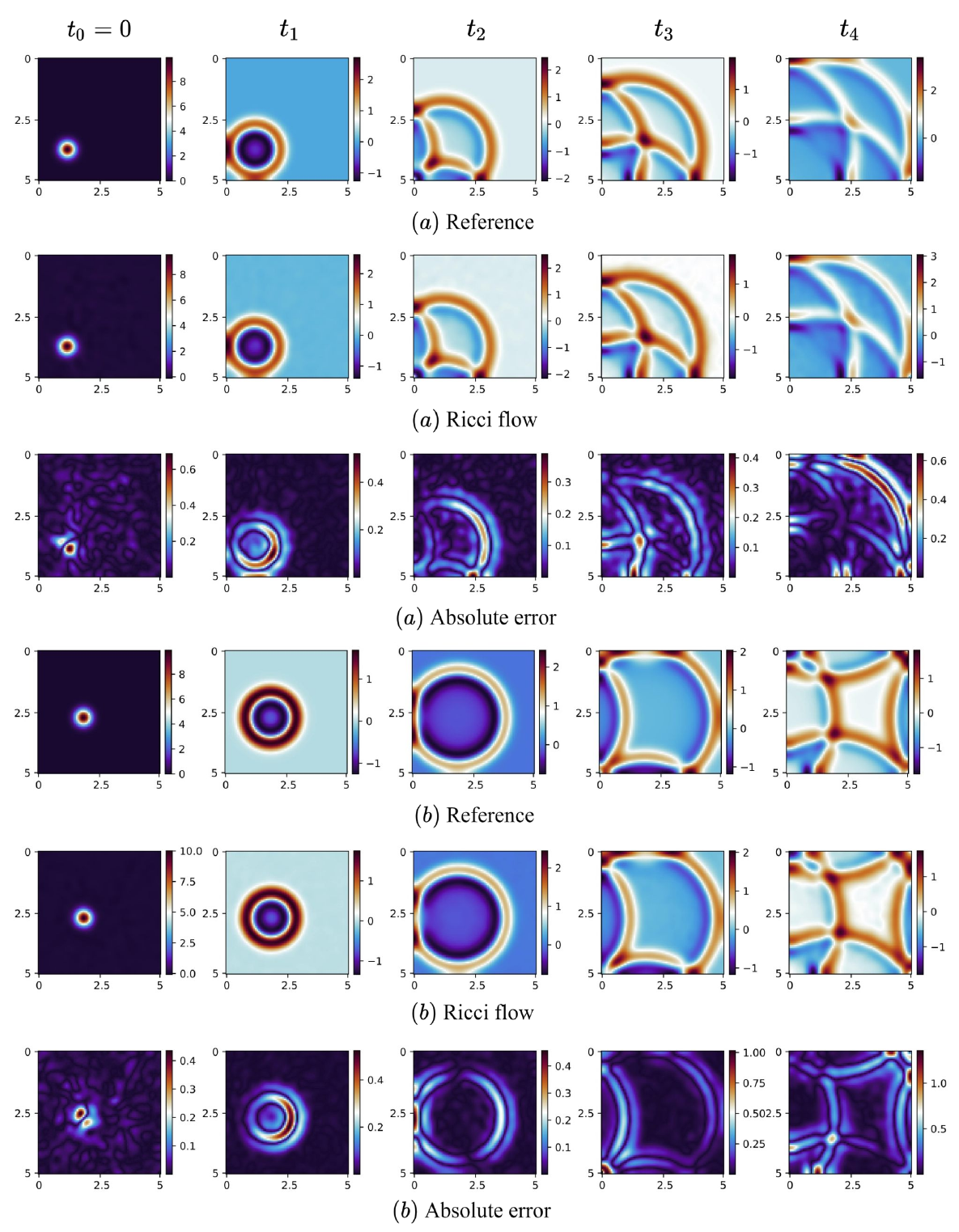}
  \caption{Comparison of numerical solution to our Ricci flow solution in the test setting on the 2-d wave equation. (a) and (b) represent two unique solutions.}
  \label{fig:2d_wave_eq_examples}
\end{figure*}

As a supplemental experiment, we deploy our Ricci flow-based method on learning the 2-d wave equation. A finite difference numerical scheme was used for constructing datasets for evaluation \citep{2dwavegithub}. Neumann boundary conditions were used. A domain $\mathcal{X} = [0,5]\times[0,5]$ was used for space, with time in $[0,4]$. Initial conditions were generated using a random Gaussian impulse of the form $\phi_0 = 10 \exp \{ -\frac{(x-\mu_1)^2}{0.1} -\frac{(y - \mu_2)^2}{0.1}\}$, where $(\mu_1,\mu_2) \in U([1,4]\times[1,4])$ is uniformly random. $1,000$ initial condition samples were used with $100$ times taken for each initial sample for a training dataset of $1,000 \times 100 = 100,000$ solutions.

\vspace{2mm}

Training details were similar to that of the Navier-Stokes experiment of section~\ref{Navier_Stokes_section}. A convolutional neural network is taken for $u_{\theta_{u}}$, mapping to $\mathcal{U} \subseteq \mathbb{R}^{100}$. The embedding is in $\mathbb{R}^{101}$. The special case of Ricci flow along the $(d-1)$-sphere is employed. Unlike the Navier-Stokes experiment, overfitting techniques were unnecessary to achieve low empirical error in the test setting, where test data lies from the same underlying distribution as that of training. In particular, injections of noise into various stages of the neural network composition, and dropout, were not needed to achieve generalization in the testing phase. As with the Navier-Stokes experiment, a deep artificial neural network had performance gains over a deconvolutional neural network, taking an architecture of $101-400-1600-4600-4600-\text{(output)}$, where output is a $80 \times 80$ mesh of the PDE solution. The final output vector is unflattened to yield the desired grid.

\section{Extrapolation sets}
\label{appendix_extrap_sets}

We list the extrapolation set used in the Burger's experiment. It is as follows:
\begin{align}
& \mathcal{B}_{\text{Burger's}}^1 = \{ \alpha \sin(2\pi x)  + \beta \cos(2 \pi x) : \alpha, \beta \in [-1,1] \}  \cup \{ (\alpha - 3 \xi^1) \sin(2\pi x) + (\beta - 3 \xi^2) \cos^3 (2 \pi x) : \\
& \ \ \ \ \ \ \ \ \ \ \ \ \ \ \ \ \ \ \ \ \alpha, \beta \in [1,2], \xi^{1,2} \sim \text{Bernoulli}(0.5), \xi^1 + \xi^2 > 0 \} ,
\end{align}
where an individual function is chosen from either set.

\vspace{2mm}

We list the extrapolation sets used in the diffusion-reaction experiment. They are as follows:
\begingroup
\allowdisplaybreaks
\begin{align}
& \mathcal{B}_{\text{diffusion}}^{1}  = \{ \alpha \sin( 2 \pi x) + (\gamma + 0.5) \cos (4 \pi x)  + \beta \sin( 4 \pi x) :  \alpha, \beta \in [-1.75, 1.75], \gamma \in [-1,1] \}  ,
\\
&  \mathcal{B}_{\text{diffusion}}^{2}  = \{ \alpha \sin(2 \pi x) + \frac{\alpha + 0.5}{2} \cos( \frac{7}{2} \pi x)  + \frac{\beta}{3} \sin( \frac{7}{2} \pi x) : \alpha, \beta \in [0,1] \} ,
\\
& \mathcal{B}_{\text{diffusion}}^{3}  = \{ \alpha \sin(2 \pi x) + \frac{\alpha + 1}{2} \cos( \frac{9}{2} \pi x)  + \frac{\beta}{3} \sin( \frac{9}{2} \pi x) : \alpha, \beta \in [0,1] \} .
\end{align}
\endgroup
Data is normalized by $C = \int_{\mathcal{X}} | \phi_0 | dx$ and scaled by $C$ for all $t \in [0,1]$ in this experiment.

\section{Training algorithm}

\vspace{-2mm}
\begin{algorithm}[htbp]
\caption{Training algorithm}\label{alg:two}
\textbf{Input:} Initial conditions $\{\tilde{\phi}_0^i\}_i$, solutions $\{ \tilde{\phi}_{t_j}^i \}_{i,j}$ evaluated at $(x, t_j) \in \Omega \times [0,T]$ \\
\textbf{Setup:} Sample  $\hat{\tau}_i = C_T t_i$, sample $\tilde{\tau}_i \in U([0,\tau])$
\begin{algorithmic}[1]
\While{$\mathcal{L}(\theta)$ is not converged}
\State Sample $n$ initial conditions $\{\tilde{\phi}_0^i\}_{i \in \mathcal{I}_n}$ and $n$ solutions $\{\tilde{\phi}_{t_j}^i\}_{i \in \mathcal{I}_n, j \in \mathcal{J}_n}$
\State $u_i = u_{\theta_{u}}(\tilde{\phi}_0^i) \in \mathcal{U}$
\State Compute encoded point on manifold $\mathcal{E}_{\theta_{\mathcal{E}}}(u_i, \hat{\tau}_j)$
\State Compute decoded solution $\mathcal{D}_{\theta_{\mathcal{D}}}(\mathcal{E}_{\theta_{\mathcal{E}}}(u_i, \hat{\tau}_j))$
\State Compute {\color{Cerulean} \textbf{Ricci flow loss}}

\Algphase{{Decoder loss}}
\State Compute ${\mathcal{L}_{dec}(\theta)}$ directly from decoder and $\{\phi^i(\cdot, t_j)\}$

\Algphase{{Metric loss}} 
\State Compute $\langle \partial_l \mathcal{E}_{\theta_{\mathcal{E}}}(u_i, \hat{\tau}_j), \partial_k \mathcal{E}_{\theta_{\mathcal{E}}}(u_i, \hat{\tau}_j) \rangle$ 
\State Formulate ${\mathcal{L}_{met}(\theta)}$

\Algphase{Minimize objective.}
\State Compute ${\mathcal{L}_{Ric}(\theta)} + {\lambda_{dec} \mathcal{L}_{dec}(\theta)} + {\lambda_{met} \mathcal{L}_{met}(\theta)}$
\State Update $\theta$

\EndWhile
\end{algorithmic}
\end{algorithm}

\vspace{-2mm}
\begin{algorithm}[htbp]
\caption{{\color{Cerulean} \textbf{Ricci flow loss}}}\label{alg:two}
Compute: 
\begin{algorithmic}[1]

  \State $\partial_t g_{\theta_g}(u_i, \tilde{\tau}_i)$
  \State $g^{kl} \in G^{-1}$ 
  \State $\partial_i g_{jk}$ 
  \State $\partial_{il} g_{jk}$ 
  \State ${\Gamma_{ij}}^l =  \frac{1}{2} \sum_{k} g^{kl} ( \partial_j g_{ik} - \partial_k g_{ij} + \partial_j g_{kj} )$

  \State ${R_i}{}^{l}{}_{jk} = \partial_j {\Gamma_{ik}}^l - \partial_k {\Gamma_{ij}}^l + \sum_{p} ( {{\Gamma}_{ik}}^p {\Gamma_{pj}}^l - {\Gamma_{ij}}^p {\Gamma_{pk}}^l )  $

  \State $\text{Ric}(g(u,t))_{ik}  = \sum_{l} {{R}_i}{}^{l}{}_{lk} $

  \State Loss ${\mathcal{L}_{Ric}(\theta)}$

\end{algorithmic}
\end{algorithm}

\section{Coordinate transformations}
\label{coord_transform}

We encourage the reader to flexibly consider special cases of manifolds in which Ricci flow may be applied.
It may be of interest to transform the base coordinate system for such special case extensions. Let $(u^1,\hdots,u^{m_u}) \rightarrow (v^1,\hdots, v^{m_v}) : \mathcal{U} \rightarrow \mathcal{V}$ be a coordinate transformation from parameterization domains $\mathcal{U}$ to $\mathcal{V}$. The Christoffel symbols indeed have a closed form under a coordinate transformation, but in order to evaluate Ricci flow in the $\mathcal{V}$ based on the coordinates of $\mathcal{U}$, one only need consider the metric $g$ and Ricci tensor $\text{Ric}(g)$. Denote $\tilde{g}_{\alpha \beta}$ the metric coefficients under $\mathcal{V}$.  We have the result
\begin{equation}
\widetilde{\text{Ric}}(g)_{\alpha \gamma} =  \text{Ric}(g)_{ik} \frac{ \partial u^{i}}{\partial v^{\alpha}}  \frac{ \partial u^{k}}{\partial v^{\gamma}} .
\end{equation}

Ricci flow under a coordinate transformation holds value when considering special cases of metrics under an alternative coordinate design over Cartesian coordinates. One may be interested in a manifold defined over a domain parameterized in Cartesian coordinates, which may offer certain computational advantages, while the metric coefficients or other useful properties of the manifold are governed by an alternative coordinate system.

\vspace{2mm}

We highlight the usefulness of this on the following example. Consider the following Riemannian metric
\begin{equation}
g(u,v,t) = \begin{pmatrix} g_{11}(u,v,t) &  g_{12}(u,v,t) 
\\
g_{21}(u,v,t) & g_{22}(u,v,t)
\end{pmatrix}  = \sum_{ (u^i,u^j) \in \{u,v\} \times \{u,v\}} g_{ij}(u,v,t) du^i \otimes du^j,
\end{equation}
but fixed and not learned with a neural network. Consider a coordinate transformation such that $(u,v)\rightarrow (\tilde{u},\tilde{v})$ . Thus, the physics-informed aspect of the training objective in intrinsic dimension $2$ is
\begin{align}
\EX_t \EX_{\Phi} \Bigg[ \ \Bigg| \Bigg| \partial_t g_{\theta_g}(\tilde{u}(u(\phi_0), v(\phi_0)),\tilde{v}(u(\phi_0),v(\phi_0)),t) + 2  \begin{pmatrix} \text{Ric}_{ik} \partial_1 u^i \partial_1 u^k & \text{Ric}_{ik} \partial_1 u^i \partial_2 u^k
\\
\text{Ric}_{ik} \partial_2 u^i \partial_1 u^k & \text{Ric}_{ik} \partial_2 u^i \partial_2 u^k
\end{pmatrix} \Bigg| \Bigg|_F^2 \ \Bigg] ,
\end{align}
which is of lesser computational expense if $\text{Ric}$ is easier to calculate in the baseline coordinate system. More specifically, consider the coordinate transformation
\begin{equation}
(\tilde{u},\tilde{v}) = (e^u \cos v, e^u \sin v ) \iff (u,v) = (\log( \sqrt{ \tilde{u}^2 + \tilde{v}^2 } ), \arctan(\tilde{v}/\tilde{u} ) ),
\end{equation}
over some relevant domain $\mathcal{U}$ where the above exists (notice in general $\arctan(\tan(\cdot)) \neq \text{id}$ since invertibility requires injectivity and we did not make a domain restriction, although we remark the image of $\arctan(\cdot)$ is indeed contained within a compact set; the first term with the logarithm is always defined unless $u = -\infty$ exactly which is not mathematically sound). This coordinate transformation may be useful due to the exponential scaling, and so a domain of large measure is created from a variable of small magnitude with $u$. For example, we found adding $\text{sigmoid}$ activation to be useful in the Burgers' equation experiment, which constricts the measure of the domain. With this transformation, we can both restrict the domain to preserve nontrivial partial derivatives, but also consider an expanded domain, which amplifies the ability of the learning task. Also, it is a conformal map, and so angles but not distances are preserved, observable from the Jacobian. Consider the baseline metric of the sphere
\begin{equation}
g(u,v,t) = R(t)^2 \begin{pmatrix}  1 &  0 
\\
0 & \sin^2(u)
\end{pmatrix} .
\end{equation}
This has a simple Ricci tensor $\text{Ric}=(d-1)g$ (see Figure \ref{fig:ricci_tensor_fixed_metric}). Under a coordinate transformation, we use the formula
\begin{equation}
g_{\alpha \beta} = g_{ij} \frac{ \partial u^i}{\partial v^{\alpha}} \frac{\partial u^j}{\partial v^{\beta}} .
\end{equation}
The partials of this coordinate transformation are
\begin{equation}
J_{11} = \frac{\partial u}{\partial \tilde{u}} = \frac{\tilde{u}}{ \tilde{u}^2 + \tilde{v}^2  }, J_{12} = \frac{\partial u}{\partial \tilde{v}} =  \frac{\tilde{v}}{ \tilde{u}^2 + \tilde{v}^2 }, J_{21} = \frac{\partial v}{\partial \tilde{u}}
- \frac{ \tilde{v} }{ \tilde{u}^2 + \tilde{v}^2 }, J_{22} = \frac{\partial v}{\partial \tilde{v}} =\frac{\tilde{u}}{ \tilde{u}^2 + \tilde{v}^2 }.
\end{equation}
For brevity, we only list the metric in its $(11)$-element after the coordinate transformation, which is
\begin{align}
\label{eqn:metric_after_coord_transform}
\tilde{g}_{11} & = g_{11} \partial_{\tilde{u}} u \partial_{\tilde{u}} u + \underbrace{ g_{12} }_{ = 0 } \partial_{\tilde{u}} u \partial_{\tilde{u}} v + \underbrace{g_{21}}_{=0} \partial_{\tilde{u}} v \partial_{\tilde{u}} u + g_{22} \partial_{\tilde{u}} v \partial_{\tilde{u}} v 
\\ & = R(t)^2 \Big[ ( \frac{\tilde{u}}{ \tilde{u}^2 + \tilde{v}^2 } )^2 + \sin^2(\log(\sqrt{\tilde{u}^2 + \tilde{v}^2 })) (\frac{ \tilde{v} }{\tilde{u}^2 + \tilde{v}^2} )^2   \Big]  .
\end{align} 
Observe we can use the fact
\begin{align}
& g_{\alpha \beta} = g_{ij} \frac{ \partial u^i}{\partial v^{\alpha}} \frac{\partial u^j}{\partial v^{\beta}}, \ \ \ \ \ g_{ij } = g_{\alpha \beta} \frac{ \partial v^{\alpha}}{\partial u^i} \frac{\partial v^{\beta}}{\partial u^j} 
= g_{k l} \frac{ \partial v^{\alpha}}{\partial u^i} \frac{\partial v^{\beta}}{\partial u^j} \frac{ \partial u^k}{\partial v^{\alpha}} \frac{\partial u^l}{\partial v^{\beta}} 
= g_{k l} \delta_{ik} \delta_{jl} = g_{ij}
\end{align}
under ideal conditions to transform the metric back to its initial state. In our example, we require further application of the chain rule; it is insufficient to only substitute back in the change of variables. Since we require the entire metric under this change of variables, we omit this computation for brevity. Thus, the new metric has a more sophisticated Ricci tensor when expressed with $\tilde{u}, \tilde{v}$, as with equation \ref{eqn:metric_after_coord_transform}. Consequently, we can learn this more complicated Riemannian metric while bypassing the essential Ricci tensor calculation for this new metric, which is of significantly greater computational expense. In this particular example, the manner in which the manifold is parameterized differs. We remark a coordinate transformation does not change the intrinsic geometry.

\section{Diagonalization of the Ricci tensor}

In our experiments, it would notable to consider restricting both the metric and the Ricci tensor as diagonal, as this lowers computational expense significantly. This would also allow high-dimensional representations: for example, the computational expense of an $d \times d$ Riemannian metric would effectively be reduced to that of a sized $d$ metric. We provide the following result that this is not always meaningful. In particular, we show just because the metric is diagonal at some $t \in [0,T]$, it is not necessarily diagonal for all $\tau>t$ due to the Ricci tensor. 

\vspace{2mm}

\subsection{A vanishing Ricci off-diagonal: geometric meaning versus computational gains}

In our experiments, it would notable to consider restricting both the metric and the Ricci tensor as diagonal for computational expense purposes. We provide the following result that this is not always meaningful. In particular, we show just because the metric is diagonal at some $t \in [0,T]$, it is not necessarily diagonal for all $\tau>t$ due to the Ricci tensor. 

\vspace{2mm}

\textbf{Theorem 1 (slightly informal).} Let $g, \text{Ric} \in \Gamma(T^* \mathcal{M} \otimes T^* \mathcal{M})$ be sufficiently smooth on $[0,T], T > 0$, and let $g$ be diagonal at $t=0$. Let $g$ be subject to Ricci flow on $[0,T]$. Then $\text{Ric}$ is not necessarily diagonal for some $t$, hence neither is $g$, over $[0,T]$.

\vspace{2mm}

\textit{Proof.} Let $i \neq k$. We will use Einstein notation only if an index is repeated in lower and upper positions exactly once each. We begin by noting Hamilton's result of the time evolution of the Ricci tensor under Ricci flow \cite{hamilton1982three}
\begin{equation}
\label{eqn:ricci_tensor_time_deriv}
\partial_t \text{Ric}_{ik} = \Delta \text{Ric}_{ik} + 2g^{pr} g^{qs} R_{piqk} \text{Ric}_{rs} - 2g^{pq} \text{Ric}_{pi} \text{Ric}_{qk} .
\end{equation} 
We remark that our calculations differ by Hamilton's by notation only, but we can work with the above with our notation. First, we remark that Hamilton's Ricci tensor and ours are equivalent (see the statement following equations \ref{eqn:riemann_curv_hamilton_and_ours} and \ref{eqn:riemann_04_hamilton_ours}). With slight abuse of notation, we will denote $R_{ijkl}$ as Hamilton's notation, and the remainder as ours. Suppose necessarily that $g_{ij} = \text{Ric}_{ij} = 0$ as well as their spatial derivatives at $t=0$, i.e. the Ricci tensor is identically $0$ along the off-diagonal. We will show $\Delta \text{Ric}_{ik}$ is not necessarily $0$ due to contractions over diagonal terms, nor is $\partial_t \text{Ric}_{ik}$. First, we compute $\Delta \text{Ric}_{ik}$. First, note if the metric is diagonal, so is its inverse. Observe using the covariant derivative formulas (see \ref{eqn:cov_deriv_formula_2}, \ref{eqn:cov_deriv_formula_3}), define
\begin{align}
T_{lik} = \nabla_l \text{Ric}_{ik} = \partial_l \text{Ric}_{ik} - {\Gamma_{il}}^m \text{Ric}_{mk} - {\Gamma_{kl}}^m \text{Ric}_{im}  .
\end{align}
Therefore
\allowdisplaybreaks
\begin{align}
\nabla_j \nabla_l \text{Ric}_{ik} = \nabla_j T_{lik} = \partial_j T_{lik} - {\Gamma_{jl}}^m T_{mik} - {\Gamma_{ji}}^m T_{lmk} - {\Gamma_{jk}}^m T_{lim} .
\end{align}
We can notice
\begin{align}
& \Delta \text{Ric}_{ik} = g^{jl} \nabla_j \nabla_l \text{Ric}_{ik} = g^{jl} \nabla_j ( \partial_l \text{Ric}_{ik} - {\Gamma_{il}}^m \text{Ric}_{mk} - {\Gamma_{kl}}^m \text{Ric}_{im} )  
\\[1em]
& = \sum_j g^{jj} \Big[ \partial_j (\nabla_j \text{Ric}_{ik}) - {\Gamma_{jj}}^m \nabla_m \text{Ric}_{ik} - {\Gamma_{ji}}^m \nabla_j \text{Ric}_{mk} - {\Gamma_{jk}}^m \nabla_j \text{Ric}_{im} \Big]
\\[1em]
&= \sum_j \Bigg[ g^{jj} \Bigg( \partial_j  ( \partial_j \text{Ric}_{ik} - {\Gamma_{ij}}^m \text{Ric}_{mk}  - {\Gamma_{kj}}^m \text{Ric}_{im} )  - {\Gamma_{jj}}^m ( \partial_m \text{Ric}_{ik} - {\Gamma_{im}}^k \text{Ric}_{kk} - {\Gamma_{km}}^i \text{Ric}_{ii} ) 
\\
& \ \ \ \ \ \ \ \ \ \ \ \ - \sum_m  {\Gamma_{ji}}^m (  \partial_j \text{Ric}_{mk} - {\Gamma_{mj}}^k \text{Ric}_{kk} - {\Gamma_{kj}}^m \text{Ric}_{mm})  - \sum_m {\Gamma_{jk}}^m ( \partial_j \text{Ric}_{im} - {\Gamma_{ij}}^m \text{Ric}_{mm} -  {\Gamma_{mj}}^i \text{Ric}_{ii})  \Bigg) \Bigg]  .
\end{align}
Let us now apply our diagonal constraints. Note the symmetry in the expression $- {\Gamma_{ij}}^k \text{Ric}_{kk} - {\Gamma_{kj}}^i \text{Ric}_{ii}$. Now, we note the following: ${\Gamma_{ij}}^k = 0$ for diagonal metrics if $i \neq j \neq k$ \cite{diagonal_christoffel_symbols}. For the first term, $j=i,k$ otherwise we get triviality. For the inner summations, we must have $m = j,i,k$, and $j$ can run over any index. Hence,
\begin{align}
= & \sum_{j \in \{i,k\}} \Bigg[ g^{jj} \partial_j ( - {\Gamma_{ij}}^k \text{Ric}_{kk} - {\Gamma_{kj}}^i \text{Ric}_{ii} ) \Bigg]
\\ & + \sum_j g^{jj} \sum_m {\Gamma_{jj}}^m \Bigg[ {\Gamma_{im}}^k \text{Ric}_{kk} + {\Gamma_{km}}^i \text{Ric}_{ii} \Bigg]
\\ & + \sum_j g^{jj} \Bigg[ - {\Gamma_{ji}}^k \partial_j \text{Ric}_{kk} + \sum_{ m \in \{j,i,k\} \ \text{distinct}} \Big( {\Gamma_{ji}}^m {\Gamma_{mj}}^k \text{Ric}_{kk} + {\Gamma_{ji}}^m{\Gamma_{kj}}^m \text{Ric}_{mm} \Big) \Bigg]
\\ & + \sum_j g^{jj} \Bigg[ - {\Gamma_{jk}}^i \partial_j \text{Ric}_{ii} + \sum_{ m \in \{j,i,k\} \ \text{distinct}} \Big( {\Gamma_{jk}}^m {\Gamma_{ij}}^m \text{Ric}_{mm} + {\Gamma_{jk}}^m {\Gamma_{mj}}^i \text{Ric}_{ii} \Big) \Bigg] := (a) .
\end{align}
Now, returning to equation \ref{eqn:ricci_tensor_time_deriv}, we have
\begin{align}
\label{eqn:eqn_38}
\partial_t \text{Ric}_{ik} & = (a) + \sum_{p,q} 2 g^{pp} g^{qq} R_{piqk} \text{Ric}_{pq} - 2 g^{ik} \text{Ric}_{ii} \text{Ric}_{kk} = (a) + \sum_q 2 (g^{qq})^2 g_{qm} {{R}_k}{}^{m}{}_{qi} \text{Ric}_{qq} - 2 g^{ik} \text{Ric}_{ii} \text{Ric}_{kk}  \displaybreak[1]
\\ & = (a) + \sum_q 2 \Bigg[  (g^{qq})^2 g_{qq}  {{R}_k}{}^{q}{}_{qi}
\text{Ric}_{qq}  \Bigg] - 0 = (a) + \sum_q 2 \Bigg[ \underbrace{ g^{qq} {{R}_k}{}^{q}{}_{qi}
\text{Ric}_{qq} }_{=(b)}  \Bigg] - 0 = (c) .
\end{align}
Denoting $R$ and $\widetilde{R}$ for Hamilton's and our notation respectively, notice that 
\begin{equation}
\label{eqn:riemann_curv_hamilton_and_ours}
{R_i}{}^{l}{}_{jk} = {\widetilde{R}_k}{}^{l}{}_{ij} \ \ \ \ \ \implies \ \ \ \ \ R_{ijkl} = \widetilde{R}_{likj} ,
\end{equation}
where $\overline{A}_{ijkl} = g_{km} {\overline{A}_i}{}^{m}{}_{jl}$ for tensor $\overline{A}$. In particular,
\begin{equation}
\label{eqn:riemann_04_hamilton_ours}
R_{ijkl} = g_{km} {{R}_i}{}^{m}{}_{jl} = g_{km} {\widetilde{R}_l}{}^{m}{}_{ij} = \widetilde{R}_{likj} .
\end{equation}
We remark the diagonalization of $g$ reduces the summation over $m$ in $\ref{eqn:eqn_38}$ to a single term, which means that ${{R}_k}{}^{q}{}_{qi}$ is not Einstein notation and does not reduce to a Ricci curvature term. In general, $(c)$ is not identically $0$. Observe it is possible terms involving a product of the Riemannian tensor and Ricci curvature appear in $(a)$ and $(b)$, but we need not do this computation. The first summation in $(a)$ only runs over $j=i,k$, but the summation in $(b)$ runs over any index. This is a guarantee that not all terms necessarily cancel out (specifically for intrinsic dimension $> 2$). We provide a non-rigorous mathematical argument why the result follows in the remark below.

\vspace{2mm}

Thus, the time derivative is nonzero, so the off-diagonal of $\text{Ric}$ is not identically zero and we are done. In particular, we have
\allowdisplaybreaks
\begin{align}
g_{ij}(t) & = g_{ij}(0) -2 \int_0^t (\text{Ric}_{ij}(0) + \int_0^{\tau} \partial_{t} \text{Ric}_{ij}(\gamma) d\gamma ) d\tau    = -2 \int_0^t \int_0^{\tau} \partial_{t} \text{Ric}_{ij}(\gamma) d\gamma d\tau     
\end{align}
is not necessarily 0. We remark we also use smoothness, and so $\partial_t \text{Ric}_{ik}$ is not only nonzero on a set of measure 0.
$\square$

\vspace{2mm}

\textit{Remark.} We develop a non-rigorous qualitative argument that the condition of $(a)$ is nonzero. Note that all terms in $(a)$ can be nonzero (this is only not true in certain circumstances, such as the Christoffel symbols are constant, thus differentiation yields $0$). Thus, we can scale the metric nonlinearly, such as through $g_{ij} \rightarrow F_{ij}(g_{ij})$, if the equation is not automatically nonzero for choice of metric. In general, for the equation $x+y=0, x,y \neq 0$, we can construct a nonlinear function $f$ such that
\begin{equation}
f_1(x) + f_2(y)  \neq 0,
\end{equation}
where $f_1, f_2$ are transformations that all relate to a parent $f$ in some way, such as through the chain rule. For us, $x$ is $(a)$ and $y$ is $(c)$. We also remark our goal is to show there exists such a metric, not that the result holds for all metrics. A nonlinear scaling changes the intrinsic geometry in general.

\vspace{2mm}

\textit{Remark.} Observe the quantity $(a)$ has close connections to the Riemannian tensor if we were to differentiate the Christoffel symbols and rearrange the summations.

\vspace{2mm}

\textit{Remark.} A key of this proof is the fact that the Laplacian of the Ricci tensor along the off-diagonal also depends on the diagonal elements. This is observable with the covariant derivative formula on a $(0,2)$ tensor
\begin{equation}
\label{eqn:cov_deriv_formula_2}
\nabla_l A_{ik} = \partial_l A_{ik} - {\Gamma_{il}}^m A_{mk} - {\Gamma_{kl}}^m A_{im} ,
\end{equation}
and since there are two summations over $m$, and the covariant derivative formula on a $(0,3)$ tensor
\begin{align}
\label{eqn:cov_deriv_formula_3}
\nabla_j A_{lik} = \partial_j A_{lik} -{\Gamma_{jl}}^m A_{mik} - {\Gamma_{ji}}^m A_{lmk} - {\Gamma_{jk}}^m A_{lim} .
\end{align}
As we mentioned, we normally have ${\Gamma_{il}}^k = 0$ for diagonal metrics if $i \neq l \neq k$ \cite{diagonal_christoffel_symbols}, but in our proof, we also have summation over $l$, so we do not automatically get triviality.

\vspace{2mm}

\begin{figure*}[htbp]
  \vspace{0mm}
  \centering
  \includegraphics[scale=0.6]{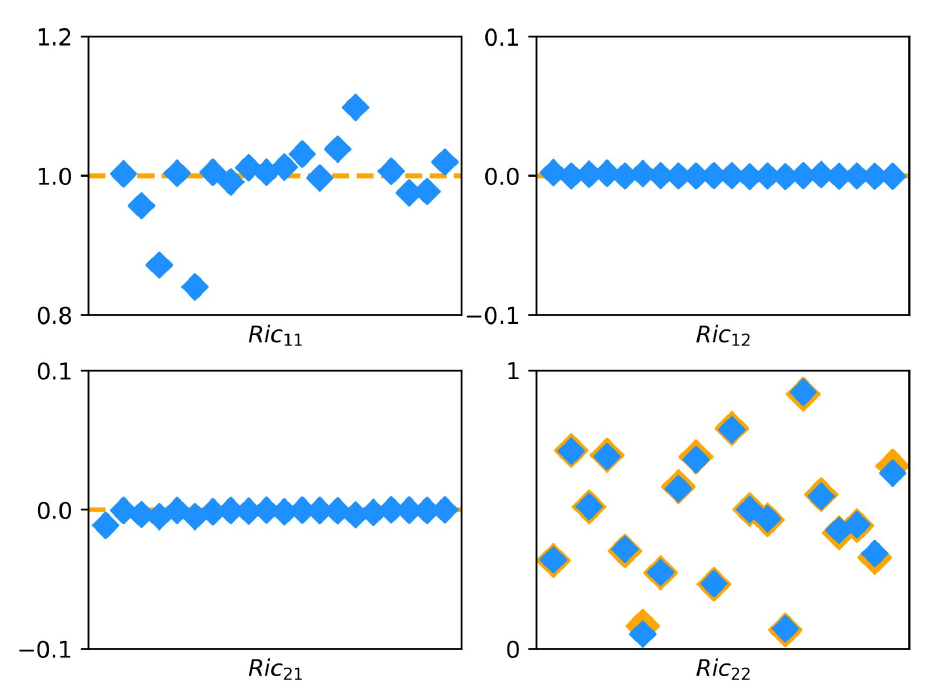}
  \caption{We demonstrate empirical results of our computed Ricci tensor (blue) compared to true Ricci tensor solutions (orange), where we learn the metric neural network $g_{\theta_g}$ upon the exact metric of the $2$-sphere. It is known the Ricci tensor for the sphere with any radius is $$\text{Ric} = (d-1)g,$$ 
  which is demonstrated in \cite{ricciflowsheridan}, where $d$ is the intrinsic dimension here. $g$ is the metric of the unit sphere. With $d=2$, we have $g = du^1 \otimes du^1 + \sin^2(u^1) du^2 \otimes du^2$ (observe the radius is not here). Training is done on $(u^1,u^2,t) \in [0,\pi]\times[0,\pi]\times[0,0.5]$ with $\text{radius} = 1$ at $t=0$. The orange lines are at $0$ for the $(12),(21)$-cases. For the $(11),(12),(21)$-cases, we evaluate the Ricci tensor within $[0.25,\pi/2]$ (we avoid $u^1 = 0$ because the metric tends to $0$ elements here and is not invertible), and $t$ is evaluated in a discretization $[0.1,0.4]$ progressively along each point in the figure. The $(22)$-case is at $u^1 \in [0.25,\pi/2-0.25]$ and $t$ is in the same discretization, and we plot the computed Ricci solution and the true $\sin^2(u^1)$. Note this experiment was conducted as Ricci tensor verification, and this experiment was not used to learn an ambient PDE. }
  \label{fig:ricci_tensor_fixed_metric}
\end{figure*}

\textit{Remark.} As a sanity check, we examine the unit $2$-sphere. We have $\text{Ric}=g$ (see Figure \ref{fig:ricci_tensor_fixed_metric}). Now, using the formulas of \cite{diagonal_christoffel_symbols},
\begin{align}
& {\Gamma_{11}}^2 = -\frac{1}{2} \frac{1}{\sin^2(u^1)} (0) = 0, {\Gamma_{21}}^1 = 0, {\Gamma_{22}}^1 = - \frac{1}{2} \partial_1 ( \sin^2(u^1) ) = -\frac{1}{2} 2 \sin(u^1) \cos(u^1) 
\\ 
& {\Gamma_{12}}^2 = \partial_1 \log( \sqrt{ \sin^2(u^1)} ) = \partial_1 \frac{1}{2} \log( \sin^2 (u^1) ) = \frac{1}{2} \frac{1}{\sin^2(u^1)} ( 2 \sin(u^1) \cos(u^1) ) = \cot(u^1) 
\\
& {\Gamma_{11}}^{1} = 0, {\Gamma_{22}}^2 = \partial_2 ( \log( \sqrt{ \sin^2(u^1) } )) = 0 .
\end{align}
Hence, using our formula of $(c)$,
\begin{equation}
\partial_t \text{Ric}_{12} = g^{11} \Big[  \partial_1 ( 0 ) +  0  + 0   \Big] + g^{22} \Big[ \partial_2 ( 0 ) + 0 + 0  \Big] + 0 + 0 = 0 ,
\end{equation}
thus our proof is not contradicted and we maintain a diagonal metric as desired, which we know to be true for a sphere subject to Ricci flow in any dimension (see section \ref{sec:sphere}).

\section{Special cases}
\label{special_cases}

Special cases can be under consideration, which offer various advantages, the primary of which is reduction in computational power necessary to conduct the learning. This permits higher dimensional structures for more expressive representations. Specifications can be made that consolidate the possibilities of the manifold into a single structure, which generally constrain the problem, but also gain computational favor in the learning stage and yield possibly desirable results, especially if the manifold accords well with the PDE data.

\vspace{2mm}

Our special cases will be constraining the metric in one that is known, i.e. inherent to an exact and known manifold, using a known solution to Ricci flow via the manifold equation itself given a domain of parameterization, and a restriction to surfaces of revolution. One can extend special cases more generally, such as in using different coordinate systems, in which the result of Appendix~\ref{coord_transform} is useful.

\subsection{Setting a known metric}
\label{known_metric}

In this setting, we specify a scenario with a known metric such that Ricci flow is satisfied. We will examine the cigar soliton, with initial metric given by
\begin{equation}
g_0 = g(u,0) = \frac{du^1 \otimes du^1 + du^2 \otimes du^2}{1 + (u^1)^2 + (u^2)^2} ,
\end{equation}
which can be solved and extended for all times under Ricci flow. The cigar soliton metric for general times can be solved as in \citep{ricciflowhenot}, which yields the metric 
\begin{equation}
g(u,t) = \frac{du^1 \otimes du^1 + du^2 \otimes du^2}{e^{4t} + (u^1)^2 + (u^2)^2} .
\end{equation}
This metric replaces the metric neural network $g_{\theta_g}$ implementation within the loss function, as well as the use of Ricci flow upon $g_{\theta_g}$ in the physics-informed setting. Now, we match the inner product of the tangent vectors upon $\mathcal{E}_{\theta_{\mathcal{E}}}$ with this fixed metric directly. Our optimization problem becomes the loss minimization framework
\begin{align}
\label{eqn:loss_cigar_soliton}
\EX \Big[  || \mathcal{D}_{\theta_{\mathcal{D}}}(\mathcal{E}_{\theta_{\mathcal{E}}}(u, \hat{\tau})) - \tilde{\phi}_{t} ||_2^2 
+ \frac{1}{2} \sum_{j=1}^2 \Big| ( e^{4 \hat{\tau}} + || u ||_2^2 )^{-1} - || \partial_j \mathcal{E}_{\theta_{\mathcal{E}}}(u, \hat{\tau}) ||_2^2 \Big|^2  +   \Big|\langle \partial_1 \mathcal{E}_{\theta_{\mathcal{E}}}(u, \hat{\tau}), \partial_2 \mathcal{E}_{\theta_{\mathcal{E}}}(u, \hat{\tau} \rangle \Big|^2   \Big] .
\end{align}

The first term is that to match the PDE solution, and the second and third are to enforce the cigar soliton manifold constraint. The loss function may be transformed for any fixed metric, generally minimizing the loss between the metric and $(J \mathcal{E}_{\theta_{\mathcal{E}}}(u, \tilde{\tau}))^T(J \mathcal{E}_{\theta_{\mathcal{E}}}(u, \tilde{\tau})) $, being the matrix of inner products of the tangent vectors, that we saw previously.

\vspace{2mm}

Another option is the torus, which has metric \citep{jain2022physicsinformed}
\begin{equation}
g(u,t) = (b + a\cos( u^2 ))^2 du^1 \otimes du^1 + a^2 du^2 \otimes du^2 
\end{equation}
over domain $u^1 \times u^2 \in [0,2\pi]\times[0,2\pi]$. We will specify $b=2, a=-1$. As with the cigar soliton, the matrix of the inner product of the tangent vectors can be matched with the fixed metric in training. A loss function similiar to~\ref{eqn:loss_cigar_soliton} can be formulated with this metric; however, we remark this metric is independent of time, and the solution is learned purely by the displacement of the manifold through the embedding rather than through the flow. 

\vspace{2mm}

One can enforce symmetry constraints upon the torus by the addition of the terms
\begin{align}
 & \EX_t \EX_{\Phi} [ ||(J \mathcal{E}_{\theta_{\mathcal{E}}}(u^1, u^2, \tilde{\tau}))^T(J \mathcal{E}_{\theta_{\mathcal{E}}}(u^1, u^2, \tilde{\tau}))  - (J \mathcal{E}_{\theta_{\mathcal{E}}}(u^1 + \delta, u^2, \tilde{\tau}))^T(J \mathcal{E}_{\theta_{\mathcal{E}}}(u^1 + \delta, u^2, \tilde{\tau}))  ||_F^2 ] 
 \\ 
 & + \EX_t \EX_{\Phi} [ || (J \mathcal{E}_{\theta_{\mathcal{E}}}(u^1, u^2, \tilde{\tau}))^T(J \mathcal{E}_{\theta_{\mathcal{E}}}(u^1, u^2, \tilde{\tau}))   - (J \mathcal{E}_{\theta_{\mathcal{E}}}(u^1, 2 \pi - u^2, \tilde{\tau}))^T(J \mathcal{E}_{\theta_{\mathcal{E}}}(u^1, 2 \pi - u^2, \tilde{\tau}))  ||_F^2]
\end{align}
to the loss, using the symmetry conditions for the torus $g(u^1, u^2, \tilde{\tau}) = g(u^1 + \delta, u^2, \tilde{\tau})$, $  g(u^1, u^2, \tilde{\tau}) = g(u^1, 2 \pi - u^2, \tilde{\tau})$ for $\delta \in [0,2\pi]$ \citep{jain2022physicsinformed}.

\vspace{2mm}

We present some computational results on the Burger's equation experiment in such settings. It is meaningful to compare to observe differences with geometry, as such acts as a means of displaying how different geometries produce different results.

\setcounter{table}{3}

\begin{table*}[ht]
\small
\centering
\begin{tabular}{wl{1.7cm} | P{1.8cm} P{1.8cm} P{1.8cm} P{1.8cm} P{1.8cm}} 
\toprule
\belowrulesepcolor{light-gray} 
\rowcolor{light-gray} \multicolumn{6}{l}  {1-d Burger's equation learning for fixed metrics} 
\\ 
\aboverulesepcolor{light-gray} 
\midrule
{method} & {$t=0$} & {$t=0.25$} & {$t=0.5$} & {$t=0.75$} & {$t=1$} 
\\ 
\midrule
\rowcolor{LightCerulean}
{Cigar}  & {$1.38 \pm 2.02$} & {$1.48 \pm 1.38$} & {$1.62 \pm 1.64$} & {$1.61 \pm 1.88$} & {$2.17 \pm 2.40$}
\\[-0.8em]
\\
\rowcolor{LightCerulean}
{Torus}  & {$1.59 \pm  0.969$} & {$2.49 \pm 1.23$} & {$2.68 \pm 1.31$} & {$2.46 \pm 1.20$} & {$3.69 \pm 2.49$}
\\ 

\bottomrule

\end{tabular}
\caption{We examine our method on the Burger's equation experiment using the cigar soliton solution and the torus. All training details and hyperparameters are kept constant as those in the original experiment aside from the manifold learning.  }

\end{table*}

\setcounter{table}{4}

\begin{table}[htbp]
\centering
\begin{tabular}{wl{2.2cm} | P{3.4cm} } 
\toprule
\belowrulesepcolor{light-gray} 
\rowcolor{light-gray} \multicolumn{2}{l}  {1-d Burger's equation extrapolation} 
\\ 
\aboverulesepcolor{light-gray} 
\midrule
{Method} & {$\mathcal{B}_{\text{burger's}}^{1}, t=0.35$} 
\\ 
\midrule
\rowcolor{LightCerulean}{Cigar}  & {$15.5 \pm 11.3$} 
\\
\\[-0.8em]
\rowcolor{LightCerulean} {Torus} & {$14.1 \pm 5.76$} 
\\ 
\bottomrule

\end{tabular}
\vspace{3mm}
\caption{We examine extrapolation results on our fixed metrics on 30 test sets on Burger's equation data. Observe the standard deviation using the torus is significantly lower.}

\end{table}

\begin{figure*}[ht]
  \vspace{0mm}
  \centering
  \includegraphics[scale=0.55]{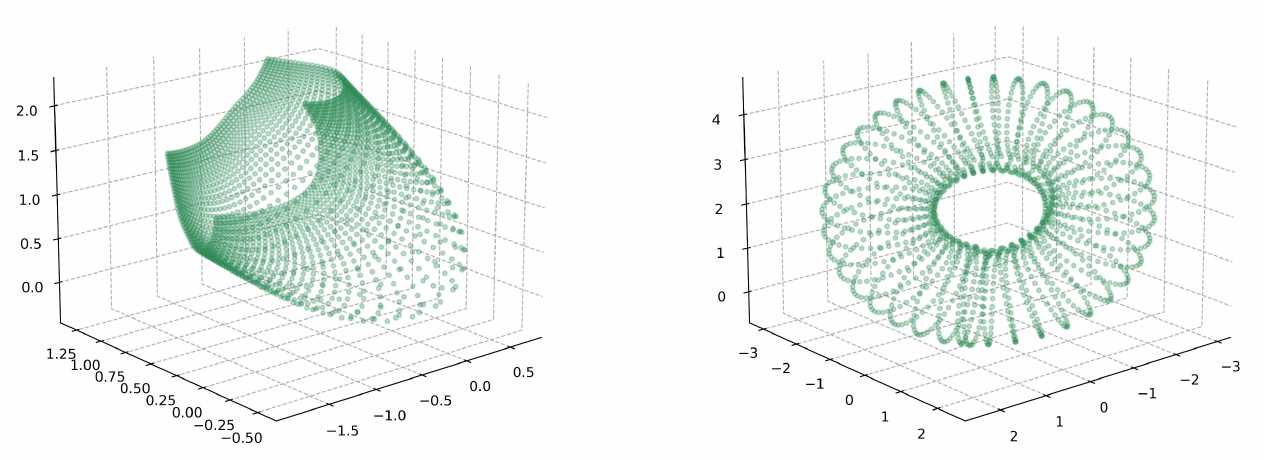}
  \caption{Examples of the cigar soliton and torus learned intrinsically in physics-informed settings. The metrics are fixed and we match the matrices of tangent vector inner products with the known metrics. This figure is purely illustrative, as the parameterization domains $\mathcal{U}$ are fixed and not learned via a neural network as they would be by learning PDE data. Only the point of the cigar is learned, but the entire torus is learned.}
  \label{fig:param_domain_burgers}
\end{figure*}

\subsection{$(d-1)$-sphere}
\label{sec:sphere}

\begin{figure*}
  \vspace{0mm}
  \centering
  \includegraphics[scale=0.63]{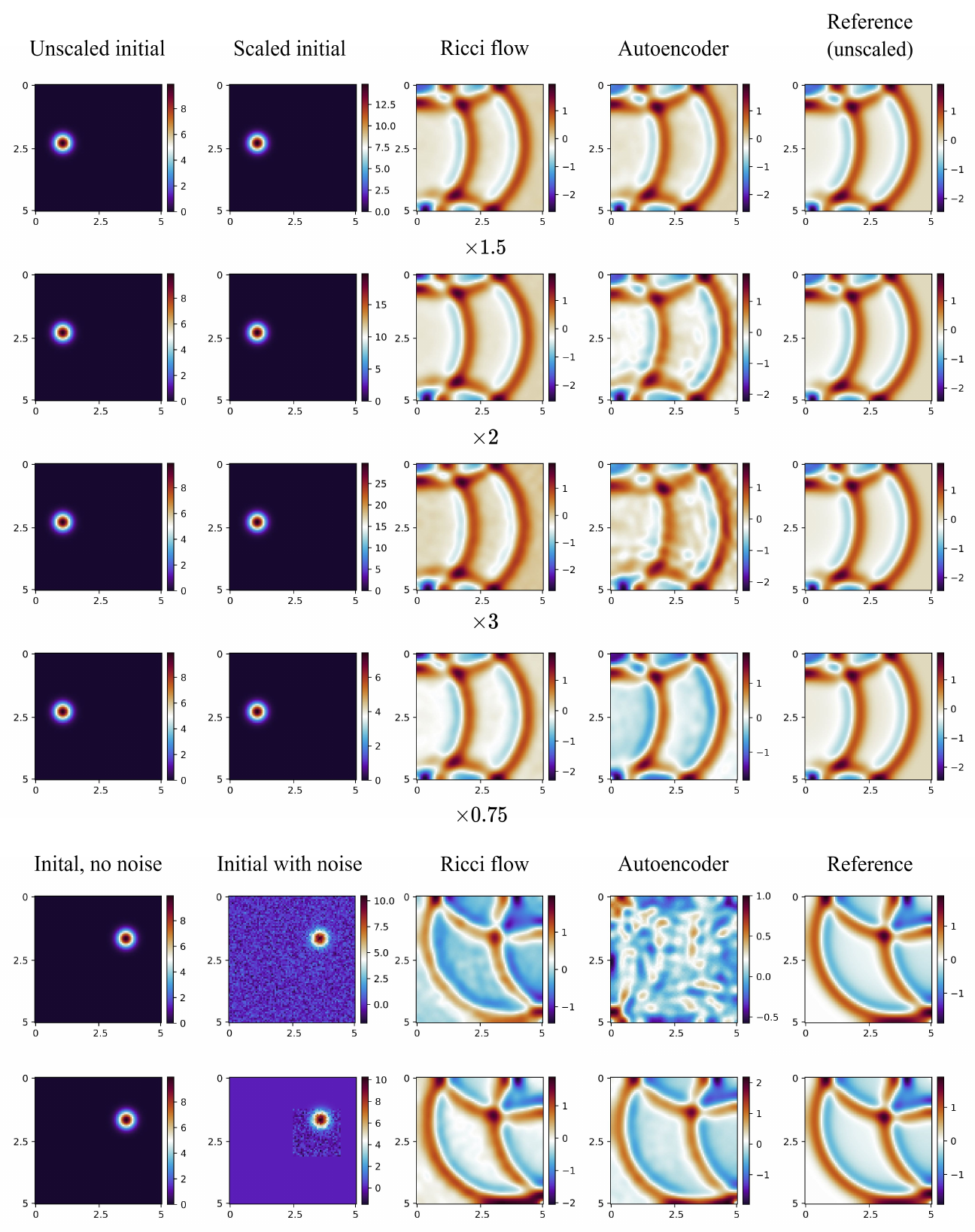}
  \caption{We present extrapolation and noise-injected robustness results on the wave equation experiment. The first four rows are on out-of-distribution data, with the initial impulse scaled by some constant. The last two are cases with noise introduced (standard deviation of $\sigma = 0.5$ for both).}
\label{fig:wave_extrap_and_noise}
\end{figure*}

A manifold evolving according to Ricci flow can be known, in which employing the metric to solve the objective function can be avoided. We will turn our attention to the $(d-1)$-dimensional sphere embedded in $\mathbb{R}^d$. The embedding exists smoothly and also isometrically in this case, but we will avoid a pullback in these experiments so this is not worth considering here. The $(d-1)$-sphere maintains that the fact that it is indeed a sphere subject to Ricci flow, while only the radius changes according to the equation
\begin{equation}
r(t) = \sqrt{ R^2 - 2(d-2) t },
\end{equation}
for initial radius $R$ \citep{ricciflowsheridan}. We will consider two cases to compute Ricci flow on the sphere. The first case is via normalization
\begin{align}
\mathcal{E}(u,\widehat{\tau}) = r(\widehat{\tau}) \cdot \frac{u}{||u||_2} ,
\end{align}
in which noise can subsequently added along this manifold. This strategy corresponds to Figure \ref{fig:navier_stokes_noise}. The second manner we also consider is how the $(d-1)$-sphere manifold parameterization is known, and given by the parameterized equations
\begin{align}
\mathcal{E}_{\mathbb{S}^{d-1}}^1 = r(t) \cos u^1, \ \ \ \ \
\mathcal{E}_{\mathbb{S}^{d-1}}^i = r(t)  \cos u^i  \prod_{j=1}^{i-1} \sin u^j, \ \ \ \ \ 
\mathcal{E}_{\mathbb{S}^{d-1}}^{d} = r(t) \prod_{j=1}^{d-1} \sin u^j ,
\end{align}
where $i \in \{2, \hdots, d-1\}$, and $(u^1, \hdots, u^{d-1}) = u \in \mathcal{U}$ is still learned via a parameterization neural network $u_{\theta_{u}}$. This strategy corresponds to Figure \ref{fig:wave_extrap_and_noise}. A decoder still maps the manifold points to the PDE solution, and so neural networks $u_{\theta_{u}}$ and $\mathcal{D}_{\theta_{\mathcal{D}}}$ remain while the metric network $g_{\theta_g}$ is removed, and the encoding network $\mathcal{E}_{\theta_{\mathcal{E}}}$ replaced.

\vspace{2mm}

Our objective function in this setting is greatly simplified, and becomes the risk minimization framework
\begin{equation}
\min_{\theta \in \Theta} \EX_t \EX_{\Phi}  [\frac{1}{N} \sum_j | \mathcal{D}_{\theta_{\mathcal{D}},j} ( \mathcal{E} ( u, \hat{\tau} )) - \tilde{\phi}_{t,j} |^2 ].
\end{equation}
The inner product term is removed, as well as the Ricci flow term, which is automatically incorporated into the evaluation involving $\mathcal{E}$. This objective function reduces the PDE-solving problem to parameterizing the initial PDE data into the optimal point along the manifold, which evolves and is subsequently decoded.

\vspace{2mm}

Formulating the sphere with this method removes the fact that the evolution is governed by an intrinsic flow. In this special case, the sphere is centered at the origin, and so the embedding is fixed. The flow can be made to behave intrinsically in one of two ways.

\vspace{2mm}

The first way is to allow the sphere to shift the center away from the origin. This allows the evolution to behave similarly to how it would during an intrinsic flow due to translation but not rotation. This can be done by constructing a new neural network $\mathcal{S}_{\theta_{\mathcal{S}}} : [0,\tau] \times \Theta_{\mathcal{S}} \rightarrow \mathbb{R}^{d}$, which displaces the sphere in the embedding based on time. This displacemenent is taken for all points along the sphere such that the center of the sphere is moved. This can be done using
\begin{equation}
\tilde{\mathcal{E}}_{\mathbb{S}^{d-1}}^i = \mathcal{E}_{\mathbb{S}^{d-1}}^i + \mathcal{S}_{\theta_{\mathcal{S}}}^i ,
\end{equation}
where $\tilde{\mathcal{E}}$ is the new sphere after the change of centers. Observe $\mathcal{E}_{\mathbb{S}^{d-1}}$ is the fixed sphere at the center and $\mathcal{S}_{\theta_{\mathcal{S}}}$ is learned in training. We denote $i$ as the $i$-th coordinate in $\mathbb{R}^{d}$.

\vspace{2mm}

The second way is to utilize the known metric of the $(d-1)$-sphere, and solve for the manifold with an encoding neural network as we did in \ref{known_metric} via the inner product and metric relation. The metric of the $(d-1)$-sphere is given as
\begin{equation}
g = r(t)^2 du^1 \otimes du^1 + r(t)^2 \sum_{i=2}^{d-1} ( \prod_{j=1}^{i-1} \sin^2 u^j ) du^i \otimes du^i .
\end{equation}
It can be noted that indeed the sphere can be isometrically embedded into $\mathbb{R}^d$, but the neural network will not learn the sphere immediately, and not every geometry can be isometrically embedded as so. Thus, the immediate neural network result is not problematic but the training process to get there is.

\subsection{Surfaces of revolution}
\label{sec:surfaces_of_revolution}

Surfaces of revolution tend to pair well with periodic data, such as instances of periodicity corresponding to circular and torus-like topologies \citep{lopez2022gdvaes}. Furthermore by adding such a restriction and designing the neural networks in such a fashion to be suitable for a surface of revolution, computational gains are to be made, like with other special cases. While Ricci flow is an intrinsic flow, a manifold being a surface of revolution indeed is an extrinsic quality, in which our intrinsic methodology becomes one that is extrinsic in this special case.

\vspace{2mm}

A surface of revolution that is time-dependent is a manifold that takes the form
\begin{align}
\label{surface_of_rev_with_time}
\hat{\mathcal{E}}(u^1, u^2, \hat{\tau}) = & (r(u^1, \hat{\tau}) \cos(u^2), r(u^1, \hat{\tau}) \sin(u^2), z(u^1, \hat{\tau})) .
\end{align}
In this framework, we bypass the need to parameterize an encoder and a metric, and instead we parameterize two neural networks
\begin{align}
r_{\theta_r} & = r : \mathcal{U}^1 \times [0,\tau] \times \Theta_r \rightarrow \mathbb{R}^+ , \\
z_{\theta_z} & = z : \mathcal{U}^1 \times [0,\tau] \times \Theta_z \rightarrow \mathbb{R} ,
\end{align}
where $\mathcal{U}^1 \times \mathcal{U}^2 = \mathcal{U} \subseteq \mathbb{R}^2$ is the parameterization domain, and $\Theta_r, \Theta_z$ are finite-dimensional parameter spaces.

\vspace{2mm}

A result proves that Ricci flow preserves symmetries found in the original metric $g_{0}$ and manifold $\mathcal{M}_0$ \citep{calegariricciflow}, meaning that a surface of revolution subject to Ricci flow maintains the fact that it indeed remains a surface of revolution throughout its evolution, i.e. symmetry is preserved. Such results in the fact that the form of equation~\ref{surface_of_rev_with_time} holds over its evolution. Furthermore, a surface of revolution has a fixed metric that is consequently known over this period, which is given by \citep{elemofdiffgeo}
\begin{align}
g & = [ (\partial_{u^1} r_{\theta_r}(u^1, \hat{\tau}))^2 + (\partial_{u^1} z_{\theta_z}(u^1, \hat{\tau}))^2] du^1 \otimes du^1  + r_{\theta_r}(u^1, \hat{\tau})^2 du^2 \otimes du^2 ,
\end{align}
using the neural networks in the metric. To solve Ricci flow, one need evaluate the Ricci tensor, which can be formulated from the Christoffel symbols which have a closed form under such a metric \citep{christoffelsymbrev}:
\begin{align}
{\Gamma_{11}}^1 & = \frac{ ( \partial_{u^1} r_{\theta_r} )( \partial_{u^1}^2 r_{\theta_r} ) + ( \partial_{u^1} z_{\theta_z} )( \partial_{u^1}^2 z_{\theta_z} ) }{ ( \partial_{u^1} r_{\theta_r} )^2 + (\partial_{u^1} z_{\theta_z})^2 }, \\
{\Gamma_{22}}^1  & = - \frac{ r_{\theta_r} \partial_{u^1} r_{\theta_r} }{ ( \partial_{u^1} r_{\theta_r} )^2 + (\partial_{u^1} z_{\theta_z})^2  } , 
\\
{\Gamma_{12}}^2 & = {\Gamma_{21}}^2 = \frac{ \partial_{u^1} r_{\theta_r}}{ r_{\theta_r}}, \\  {\Gamma_{11}}^2 & = {\Gamma_{12}}^1 = {\Gamma_{21}}^1 = {\Gamma_{22}}^2 = 0 .
\end{align}
Our training objective function is transformed into 
\begin{align}
\min_{\theta \in \Theta}  \EX_{t \sim U[0,T]} \EX_{\phi_0 \sim \Phi} \Big[ & | \partial_t (\partial_{u^1} r_{\theta_r}(u^1, \hat{\tau}))^2  + \partial_t (\partial_{u^1} z_{\theta_z}(u^1, \hat{\tau}))^2 + 2 (\text{Ric}(g))_{11} |^2  \\
& +  |\partial_t r_{\theta_r}(u^1, \hat{\tau})^2 + 2 (\text{Ric}(g))_{22} |^2 
 +  |2 (\text{Ric}(g))_{12}|^2 + |2 (\text{Ric}(g))_{21}|^2
\\
& + \frac{1}{N} \sum_j | \mathcal{D}_{\theta_{\mathcal{D}},j} ( r_{\theta_r}(u^1, \hat{\tau} ) \cos(u^2), r_{\theta_r}(u^1, \hat{\tau}) \sin(u^2), z_{\theta_z}(u^1, \hat{\tau})  ) - \tilde{\phi}_{t,j} |^2 \Big] .
\end{align}
To formulate the Ricci tensor, one can avoid analytic derivatives of the Christoffel symbols via automatic differentiation.

\section{Comparable methodology for intrinsic dimension 2 and extrinsic dimension 3}  

The Riemannian tensor can additionally be computed using the coefficients of the second fundamental form \citep{elemofdiffgeo} in the case that $\text{dim}(\mathcal{U})=2$, bypassing the need of Christoffel symbols. The coefficients of the second fundamental form are a collection $\{L_{ij}\}_{ij=1}^{\text{dim}(\mathcal{U})}$ defined as
\begin{equation}
L_{ij}(t) = \langle  \partial_{ij}^2 \mathcal{E}(u,t), n(t) \rangle ,
\end{equation}
where $\mathcal{E}$ is the manifold function taken with neural network $\mathcal{E} = \mathcal{E}_{\theta_{\mathcal{E}}}$, $u \in \mathcal{U}$ belongs to a parameterization domain, and $\partial_{\ell}$ denotes the partial derivative with respect to $u^{\ell}$. $n$ is the vector normal to the surface, i.e.
\begin{equation}
n(t) = \frac{ \partial_1 \mathcal{E}(u,t) \times \partial_2 \mathcal{E}(u,t) }{ || \partial_1 \mathcal{E}(u,t) \times \partial_2 \mathcal{E}(u,t) ||_2 }  
\end{equation}
when $\mathcal{U} \subseteq \mathbb{R}^2$. The Riemannian tensor can be formulated as, up to a sign convention,
\begin{equation}
{R}_{i}{}^{l}{}_{jk}  = L_{ik} {L^l}_j - L_{ij} {L^l}_k .
\end{equation}
where we use 
\begin{equation}
{L^l}_k = \sum_i L_{ik} g^{il},
\end{equation}
where the metric coefficients $g_{ij}$ can be computed directly from the manifold $\mathcal{E}$ as in equation~\ref{eqn:metric_coefficient}. Hence, we can circumvent the use of a physics-informed neural network for the metric $g_{\theta_g}$, and substitute the Riemannian tensor directly into the Ricci flow equation \citep{li2017applying} for a residual. Ricci flow with the above formulation becomes
\begin{equation}
\partial_t  \langle \partial_i \mathcal{E}, \partial_j \mathcal{E} \rangle  = -2 \sum_{l} (L_{ij} {L^l}_l - L_{il} {L^l}_j )  .
\end{equation}
where the indices $(i,j)$ correspond to that that would be of the metric. The residual to minimize in the objective function to ensure the satisfaction of Ricci flow becomes
\begin{align}
& \mathcal{L}_{Ric}(\theta) = \EX_{t \sim U[0,T]} \EX_{\phi_0 \sim \Phi}  [ \sum_{i} \sum_j | \partial_t \langle \partial_i \mathcal{E}(u, \tilde{\tau}), \partial_j \mathcal{E}(u, \tilde{\tau}) \rangle    + 2 \sum_{l} (L_{ij} {L^l}_l - L_{il} {L^l}_j ) |^2 ] ,
\end{align}
where the summation over $(i,j)$ runs over the indices corresponding to the metric, and the summation over $l$ runs over the contraction of the Ricci tensor with respect to the Riemannian tensor. The term $\mathcal{L}_{met}(\theta)$ is nonexistent in the loss minimization framework in this strategy.

\vspace{2mm}

This alternative methodology is comparable to that in which we previously established. It still maintains computational expense, distinguished by the amount of terms to be differentiated, and additionally requires a parameterization domain $\mathcal{U} \subseteq \mathbb{R}^2$; however, we remark it is worthy to realize what we formerly proposed can be reformulated in such a way.

\section{Discussion of new non-parametric geometric flows with special cases of Perelman's $\mathcal{W}$-functional} 

In this section, we develop methodology relating to Perelman's $\mathcal{W}$-functional. Since physics-informed Ricci flow is computationally nontrivial, we can develop similar yet disparate geometric flows by solving an objective derivative of Perelman's $\mathcal{W}$-functional under special cases. We specifically examine conformally flat metrics.

\vspace{2mm}

First, we make the following remark about physics-informed neural networks (PINNs). We can simulate a physics loss by minimizing a PDE residual over a manifold which has a corresponding integral formulation of the form
\begin{equation}
|| \ f \ | |_{L^1(\mathcal{M}_t \times [0,T])} \propto \int_0^T \int_{\mathcal{M}_t} |f| \tilde{\rho} dV_t dt = \int_0^T \int_{\mathcal{U}} |  f \circ \mathcal{E}(u,t) | \rho(u) \sqrt{\text{det}(g)} du dt 
\end{equation}
for suitable Radon-Nikodym derivative $\rho$ with corresponding weighting function $\tilde{\rho}$ over the manifold, and the physics loss can be evaluated discretely with a collocation procedure. In a PINN setup, $f$ is a PDE residual, but this need not be the case. Even moreso, we can simulate integration over a manifold with the same sampling procedure as a PINN by utilizing Monte-Carlo integration, i.e.
\begin{equation}
\lim_{N \rightarrow \infty} \frac{ \int_0^T \int_{\mathcal{U}} \sqrt{\text{det}(g(u,t))} du dt }{N} \sum_{u_i \in U(\rho), 1 \leq i \leq N} |f(\mathcal{E}(u_i,t_i))| \sqrt{\text{det}(g(u_i, t_i))} \overset{P}{\rightarrow} \int_0^T \int_{\mathcal{M}_t}  |f(x)| \tilde{\rho} dV_t dt 
\end{equation}
by the (weak) law of large numbers, and we have smooth manifold map $\mathcal{E} : \mathcal{U} \rightarrow \mathcal{M}_t$ and function over the manifold $f : \mathcal{M}_t \rightarrow \mathbb{R}^d$. In practice, $\int_0^T \int_{\mathcal{U}} \sqrt{\text{det}(g(u,t))} du dt$ is constant and irrelevant for machine learning purposes. 

\vspace{2mm}

Perelman introduced the $\mathcal{W}$-functional \cite{perelman2002entropyformularicciflow} \cite{toppingricciflow}
\begin{equation}
\label{eqn:perelman_functional}
\mathcal{W}(f,g,\tau) = \int [ \tau ( R + || \nabla f ||^2) + f - n ] u dV
\end{equation}
for function $f$, metric $g$, scale parameter $\tau$, and $u = (4 \pi \tau)^{-n/2} e^{-f}$. Here, $R$ is scalar curvature. Under Ricci flow as well as other conditions, it can be shown Ricci flow is a gradient flow and that $\frac{d}{dt} \mathcal{W} \geq 0$ \cite{toppingricciflow}; however, by enforcing the functional as non-decreasing in time, we do not immediately have a manifold under Ricci flow is implied, i.e. monotonicity is not sufficient without additional conditions.

\vspace{2mm}

We propose an alternative to solving physics-informed Ricci flow in high dimensions. Even for high-dimensional diagonal metrics, computing the Ricci tensor can still be nontrivial. Instead, computing the scalar curvature, which is often derivative of the Ricci tensor, can have closed forms and can increase computational efficiency due to simplicity and backpropagation and automatic differentiation through scalars instead of higher-dimensional representations. Consequently, we propose considering the conformally flat metric
\begin{equation}
g(u,t) = e^{ 2 \varphi(u,t) } \cdot g_{\text{Euclidean}} = e^{ 2 \varphi(u,t) } \cdot I 
\end{equation}
with special case of scalar curvature \cite{wikipedia_scalar_curvature}
\begin{equation}
R = \frac{n-1}{ e^{ 2 \varphi } } \Big( -2 \Delta \varphi - (n-2)  || \nabla \varphi ||^2\Big).
\end{equation}
Here, $\Delta$ is both the Laplacian the Laplace-Beltrami operator since the original metric is flat. Thus, we propose setting $f$ as fixed and using this closed form scalar curvature in the functional of equation \ref{eqn:perelman_functional} and simulating the geometric flow, which is not necessarily Ricci flow, in a Monte-Carlo integration type objective. We demonstrated this is closely related to a physics-informed objective but not quite the same since we are not solving the PDE: we do not solve a PDE of the form
\begin{equation}
g_t = \Lambda ,
\end{equation}
but solving the functional directly solves the geometric flow for us. We also remark the above ODE is partially meaningless without additional constraints because it can be solved by learning $g = \Lambda = 0$. We can set $f = 0, \tau=1$ (or choose $f,\tau$ as something else if desired). We solve
\begin{equation}
\frac{d}{dt}\mathcal{W}_{\text{modified}}  \geq 0 
\end{equation}
with an objective of the form
\begin{align}
\text{Objective} & \propto \gamma_{\text{entropy}} \text{relu} \Bigg( \text{positive const} -  \frac{d}{dt} \mathcal{W}_{\text{modified}}(0,g,1) \Bigg) = \gamma_{\text{entropy}} \text{relu} \Bigg( 
 \text{positive const} -    \frac{d}{dt} \int_{\mathcal{M}_t} R \tilde{\rho} dV_t  \Bigg)
\\
& = \gamma_{\text{entropy}} \text{relu} \Bigg( \text{positive const} -  \frac{d}{dt} \int_{\mathcal{U}} \frac{n-1}{e^{ 2 \varphi }} \Big( -2 \Delta \varphi - (n-2)  || \nabla \varphi ||^2 \Big) \sqrt{\text{det}(g)} \rho(u) du  \Bigg)
\end{align}
with a Monte-Carlo integration-type loss, where $\varphi = \varphi_{\theta_{\varphi}}$ is a neural network, and $\gamma_{\text{entropy}} \in \mathbb{R}^+$ is a scaling coefficient towards gradient contribution. The time derivative can be taken with automatic differentiation, and we assume exchange of differentiation and integration is valid. Although this objective forces the metric to undergo evolution (observe $\varphi=\text{constant}$ has loss $\gg 0$), we do not have a closed form for the evolution, thus we describe the corresponding $g$ as non-parametric. We remark we choose $f=0$, but $u$ and $\tilde{\rho}$ actually have some relation, and so we simplify the above. Lastly, we also remark knowing $\rho$ is irrelevant, as this weighting matches uniform sampling with respect to training data $\phi_0 \sim \Phi$, as discussed in section \ref{sec:methods}.

\section{Augmented architecture}
\label{MMLP}

\begin{figure}[htbp]
  \vspace{0mm}
  \centering
  \includegraphics[scale=0.525]{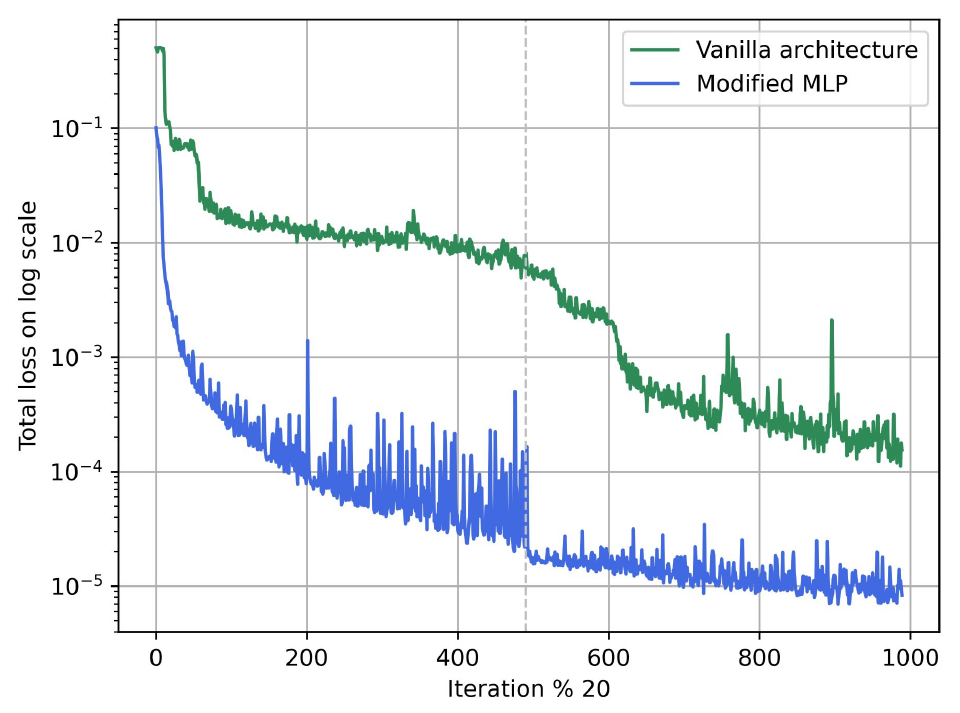}
  \caption{We compare training loss with the modified MLP architecture versus a vanilla architecture used in the diffusion-reaction experiment. The learning rate was adjusted after 10,000 iterations (dashed gray line). The first handful of iterations are omitted to better demonstrate the overall training.}
  \label{fig:loss_comparison_fig}
\end{figure}

In this section, we describe the modified multilayer perceptron architecture as in \citep{wang2021learning} \citep{wang2020understanding} for our diffusion-reaction experiment. Denote $\sigma(\cdot)$ a twice continuously differentiable activation function, and denote $\zeta$ neural network input. We first pass input $\zeta$ into two layers $\hat{\zeta} = \sigma(\hat{W} \zeta + \hat{b}), \tilde{\zeta} = \sigma(\tilde{W} \zeta + \tilde{b})$. The architecture proceeds iteratively as
\begin{equation}
\zeta^{i+1} = (1 - \sigma( W^i \zeta^{i} + b^i ) ) \odot \hat{\zeta} + \sigma( W^i \zeta^{i} + b^i ) \odot \tilde{\zeta} ,
\end{equation}
where the first iterate is given by $\zeta^1 = \sigma(W^0 \zeta + b^0)$, and $\odot$ denotes element-wise multiplication. Quantities $W^i, b^i$ denote neural network weights and biases.

\section{Additional experimental details}

We assess our experiments using relative $L^1$ error upon predicted solutions and base data, using the metric
\begin{equation}
\EX_{\Phi} \Big[ \frac{ || \mathcal{D}_{\theta_{\mathcal{D}}}(\mathcal{E}_{\theta_{\mathcal{E}}}(u,\hat{\tau})) - \phi_{\hat{t}} ||_{L^1(\mathcal{X})}    }{ || \phi_{\hat{t}} ||_{L^1(\mathcal{X})}} \Big] ,
\end{equation}
at time $\hat{t}$ corresponding to Ricci flow time $\hat{\tau}$, where an empirical average is taken to approximate the expected value. This metric behaves as a percentage error.

\vspace{2mm}

One way to facilitate the training process focuses on differentiation of the Christoffel symbols, i.e. $\partial_j {\Gamma_{ik}}^l$, which is nontrivial via automatic differentiation in a recurring training procedure, and adds significant time to the training process. This can mostly be resolved by differentiating each term involved in the Christoffel symbol calculation individually, and then forming the derivatives $\partial_j {\Gamma_{ik}}^l$ afterwards. This resulted in speeding up training by approximately $\times 3$ on a T4 GPU. When constructing our derivatives numerically, we first compute all second-order derivatives of the metric and then construct the Christoffel symbols from the derivatives of these terms via product rule, i.e.
\begin{align} \partial_j {\Gamma_{ik}}^l = & \frac{1}{2}  (\partial_j g^{pl} ( \partial_k g_{ip} - \partial_p g_{ik} + \partial_i g_{pk} )  + g^{pl} ( \partial_{kj} g_{ip} - \partial_{pj} g_{ik} + \partial_{ij} g_{pk} ) ).
\end{align}

\section{Comments on embedding dimensions}
\label{sec:embedding_dims}

Observe many of our experiments involving embedding the manifold from $\mathbb{R}^{2}$ into $\mathbb{R}^3$. Generally, with our methods, we operate under the conditions so that the embedding is defined for a manifold map from $\mathcal{U} \subseteq \mathbb{R}^{d-1} \rightarrow \mathcal{M} \subseteq \mathbb{R}^d$, such as smoothness (neural networks are generally $C^{\infty}$ with $C^{\infty}$ activations), injectivity (otherwise two distinct initial conditions have the same PDE solutions, violating well-posedness,  which we typically have); our induced metrics are full rank; and a compact domain. In our experiments with the sphere, we also consider an embedding from $\mathbb{R}^{d-1} \rightarrow \mathbb{R}^d$. The allowance of this embedding for the sphere is a well known result, thus this mapping is valid for us (see \cite{whitney_embedding}). Empirically, we did not find the existence conditions such as non-orientability in our manifolds, and properties that allow a well-defined embedding to $\mathbb{R}^d$ are generally assumed.

\onecolumn

\section{Empirical Ricci tensor evaluation}

\section{Alternative manifold views}
\label{app:burgers_eq_manifold_app}

\begin{figure*}[p]
  \centering

  \includegraphics[width=0.8\textwidth]{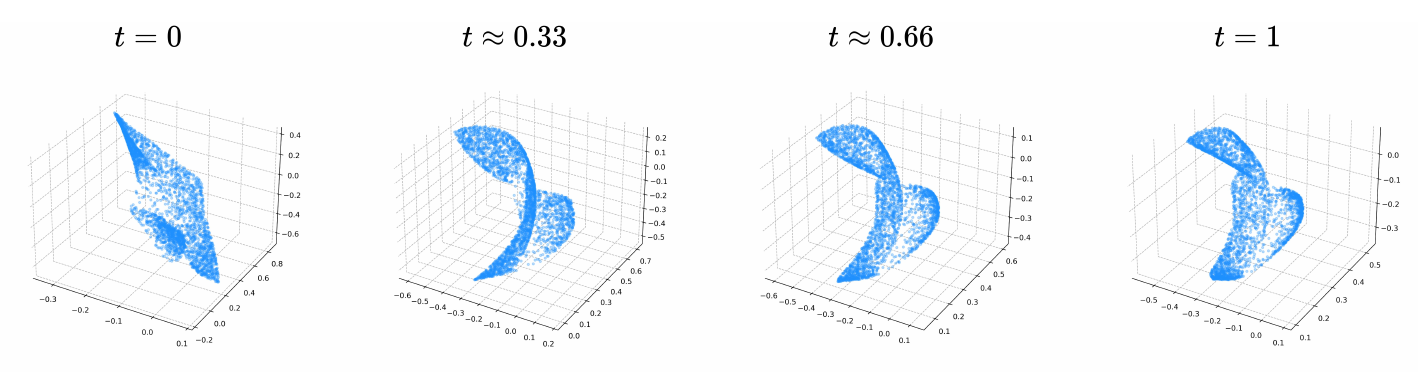}
  \caption{We provide an alternative view of the manifold as in the Burger's equation experiment.}
  \label{fig:ricci_flow_burgers_manifold_appendix}

  \vspace{0.2cm}

  \includegraphics[width=0.8\textwidth]{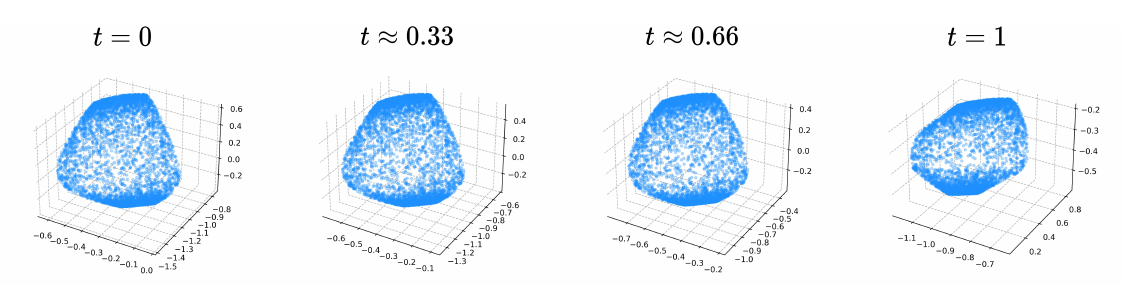}
  \caption{We provide an alternative view of the manifold as in the diffusion-reaction equation experiment.}
  \label{fig:diffreac_manifold_appendix}

  \vspace{0.2cm}

  \includegraphics[width=0.8\textwidth]{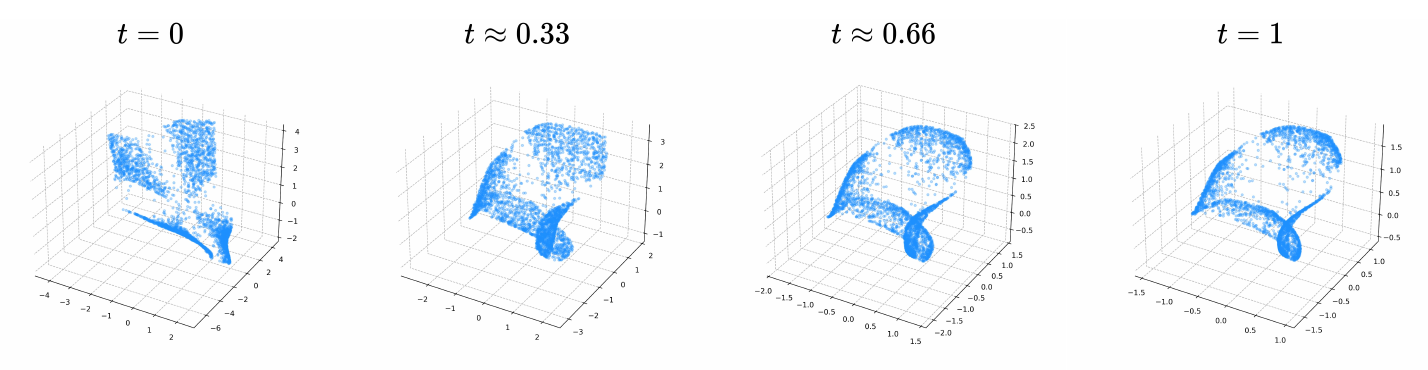}
  \caption{We provide a view of the manifold as in the extended with no regularization experiment. We use Burger's equation data here.}
  \label{fig:unreg_manifold_appendix}

  \vspace{0.2cm}

  \includegraphics[width=0.8\textwidth]{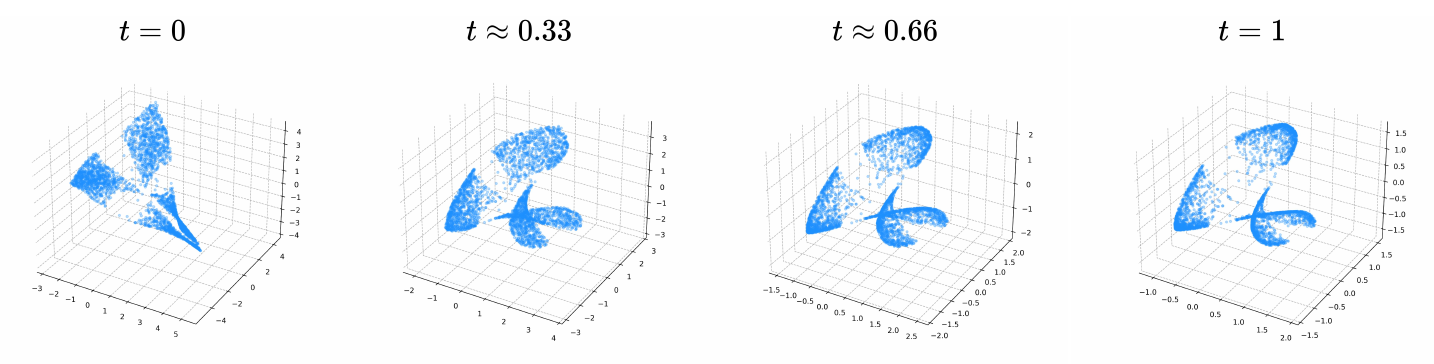}
  \caption{We provide a view of the manifold as in the extended with no regularization experiment. This figure is a reiteration of Figure \ref{fig:unreg_manifold_appendix}. It can be noted the manifold that forms is not unique, and varies upon retraining. As we can see, the metric is less canonical with respect to a pullback uniformized in curvature, and is overall a more dissonant representation.}
  \label{fig:unreg_2_manifold_appendix}
\end{figure*}

\onecolumn

\section{Additional figures}
\label{appendix_manifold_figures}

\subsection{Burger's equation}

\begin{figure*}[ht]
  \vspace{0mm}
  \centering
  \includegraphics[scale=0.775]{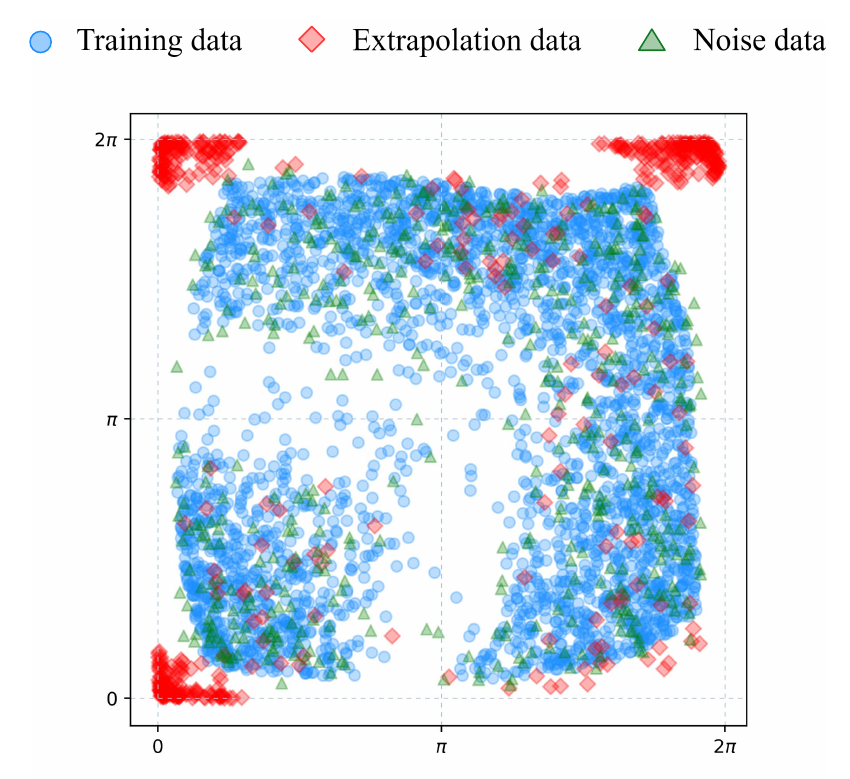}
  \caption{We view training, extrapolation, and noise data as points in the local coordinates domain $\mathcal{U}$.}
  \label{fig:param_domain_burgers}
\end{figure*}

\begin{figure*}[ht]
  \vspace{0mm}
  \centering
  \includegraphics[scale=0.625]{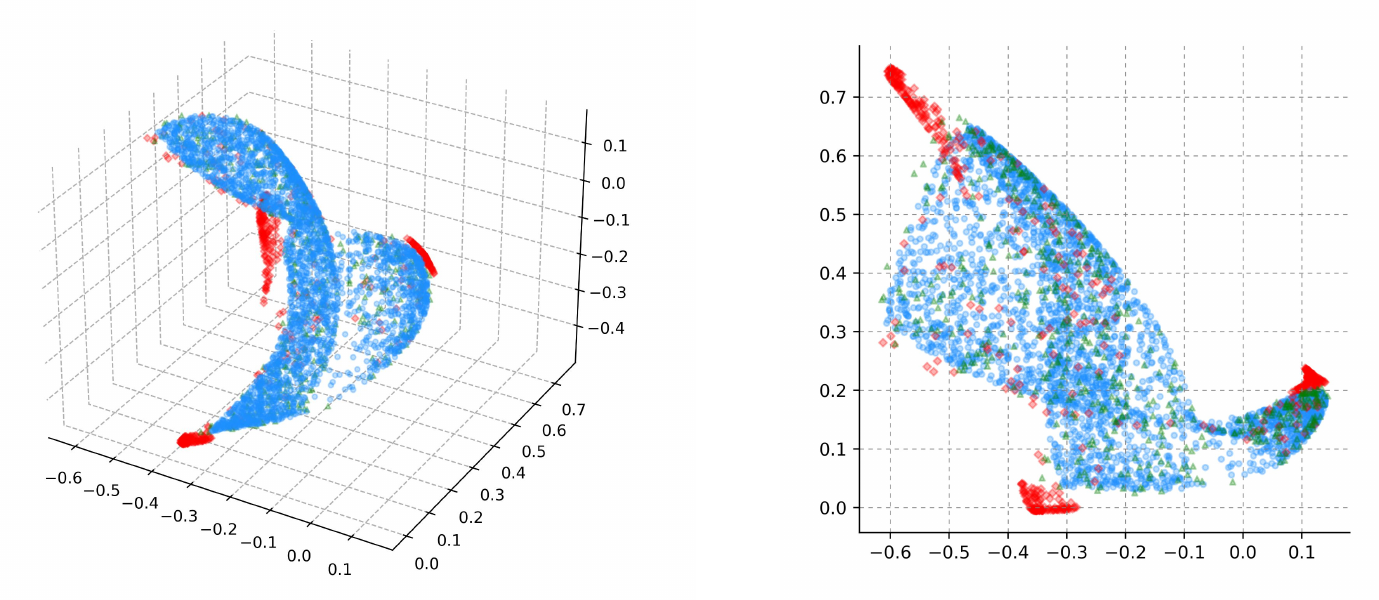}
  \caption{We view training, extrapolation, and noise data as points along the manifold at $t=0.5$. The right image is a projection on the $xy$-axis.}
  \label{fig:param_domain_burgers}
\end{figure*}

\newpage 

\subsection{Diffusion-reaction equation}

\begin{figure*}[ht]
  \vspace{0mm}
  \centering
  \includegraphics[scale=0.775]{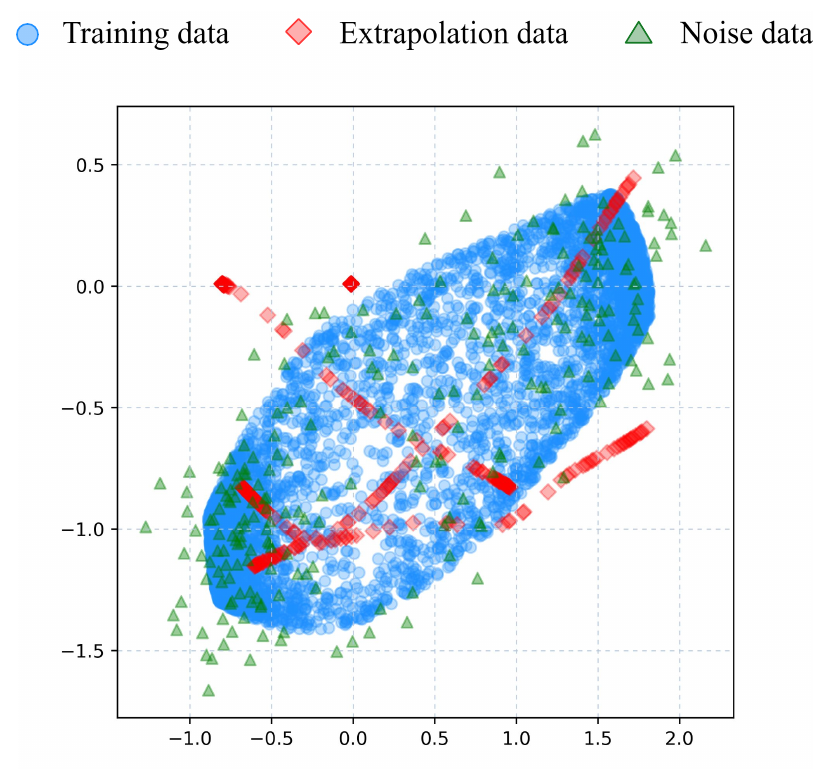}
  \caption{We view training, extrapolation, and noise data as points in the local coordinates domain $\mathcal{U}$.}
  \label{fig:param_domain_diffusion}
\end{figure*}

\begin{figure*}[ht]
  \vspace{0mm}
  \centering
  \includegraphics[scale=0.625]{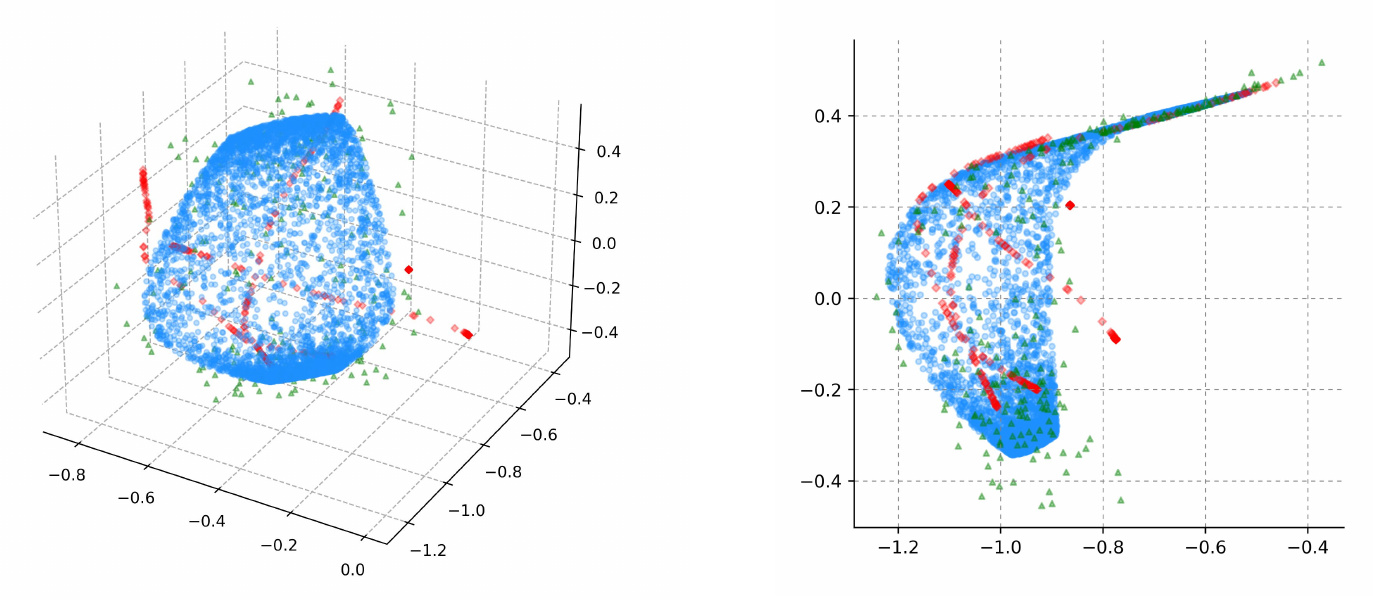}
  \caption{We view training, extrapolation, and noise data as points along the manifold at $t=0.5$. The right image is a projection on the $yz$-axis.}
  \label{fig:param_domain_diffusion}
\end{figure*}

\newpage

\subsection{Extended AE latent space with Burger's data}

\begin{figure*}[ht]
  \vspace{0mm}
  \centering
  \includegraphics[scale=0.75]{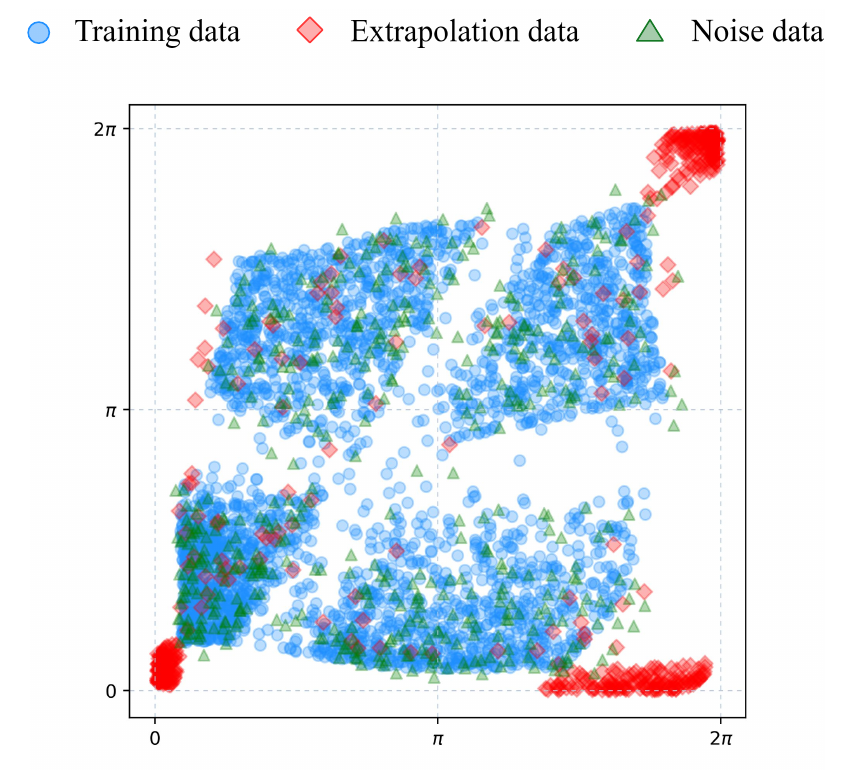}
  \caption{We view training, extrapolation, and noise data as points in the local coordinates domain $\mathcal{U}$. Here, there is no Ricci flow regularization, corresponding to the extended case as in section \ref{sec:methods}.}
  \label{fig:param_domain_noreg}
\end{figure*}

\begin{figure*}[ht]
  \vspace{0mm}
  \centering
  \includegraphics[scale=0.625]{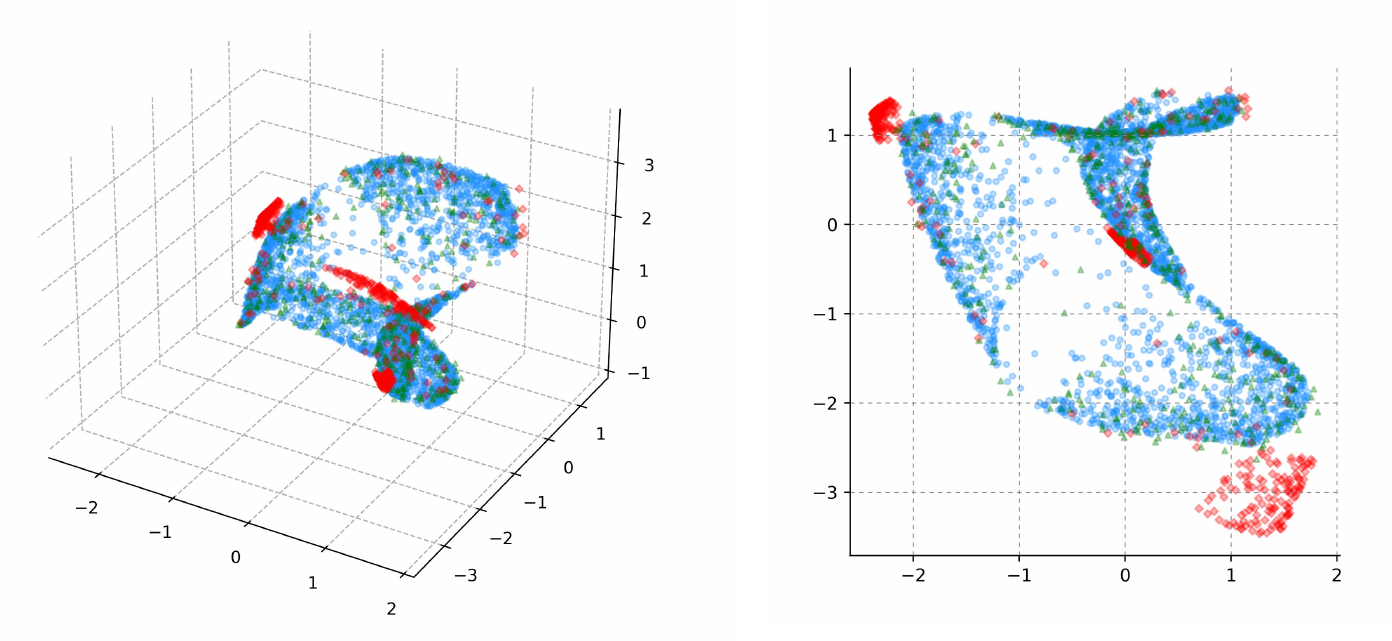}
  \caption{We view training, extrapolation, and noise data as points along the manifold at $t=0.5$. The right image is a projection on the $xy$-axis. Again, this corresponds to the unregularized case, and the manifold that forms is not unique.}
  \label{fig:noreg_manifold}
\end{figure*}

\begin{figure*}[ht]
  \vspace{0mm}
  \centering
  \includegraphics[scale=0.7]{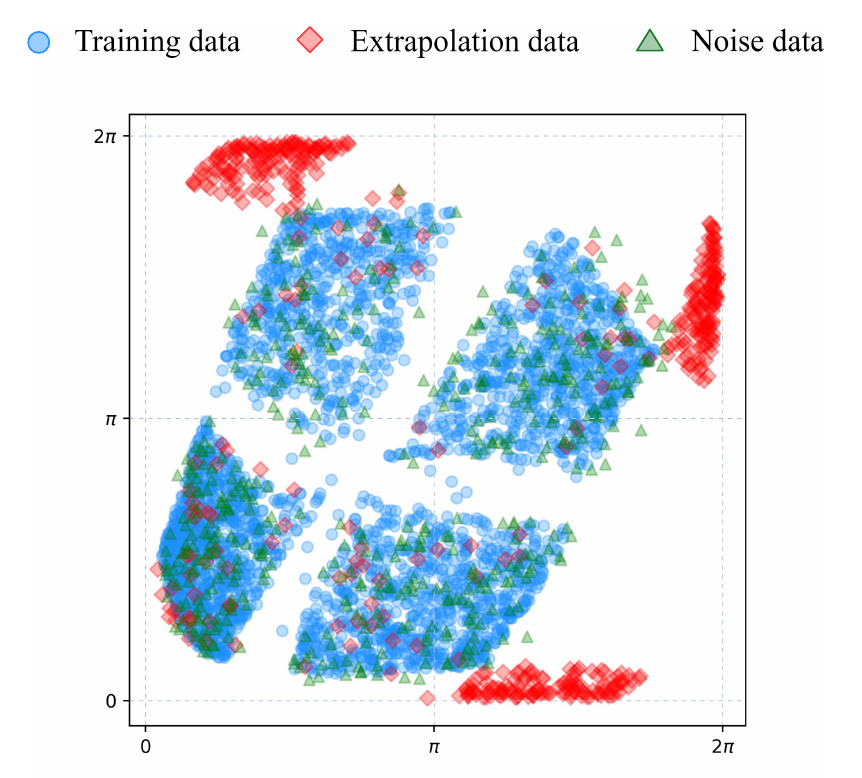}
  \caption{We view training, extrapolation, and noise data as points in the local coordinates domain $\mathcal{U}$. We plot again in the unregularized case to demonstrate the non-uniqueness of the formed latent manifold, thus this figure is consistent with Figure \ref{fig:param_domain_noreg}.}
  \label{fig:param_domain_noreg_2}
\end{figure*}

\begin{figure*}[ht]
  \vspace{0mm}
  \centering
  \includegraphics[scale=0.6]{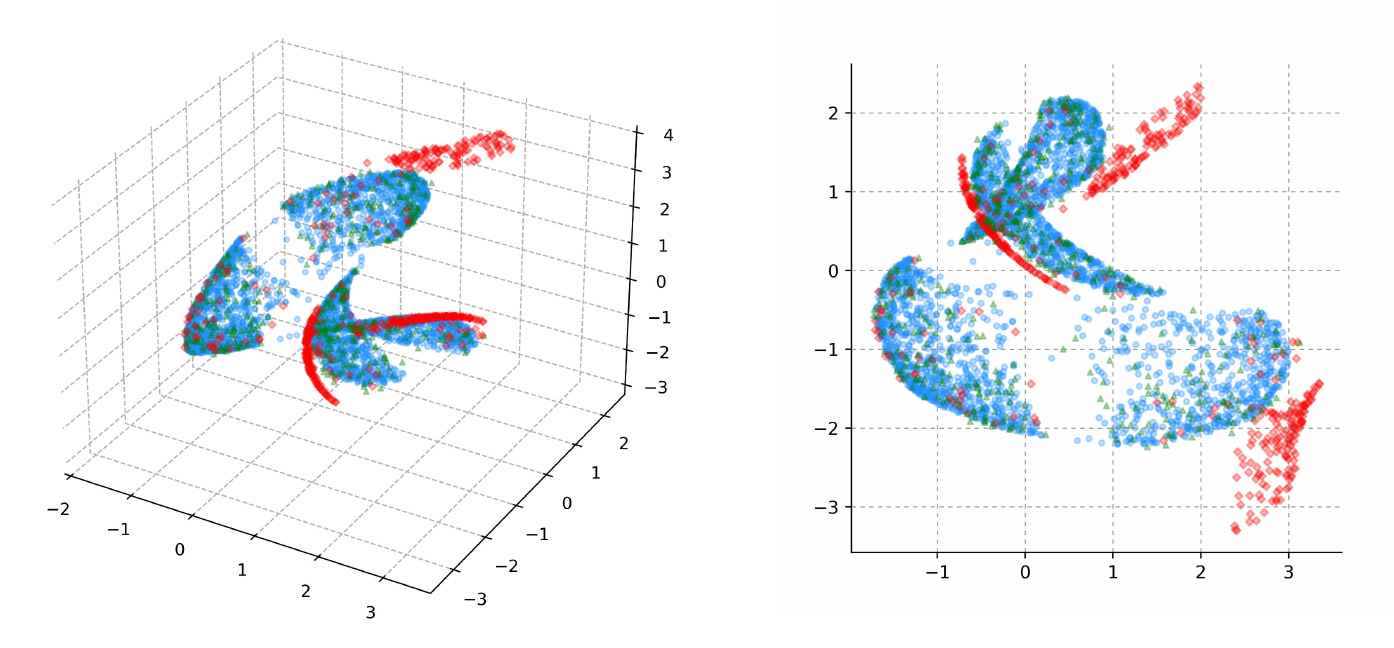}
  \caption{We view training, extrapolation, and noise data as points along the manifold at $t=0.5$. The right image is a projection on the $xy$-axis. Again, this corresponds to the unregularized case, and the manifold is not unique.}
  \label{fig:noreg_manifold_2}
\end{figure*}

\newpage

\section{Additional wave equation figure}

\end{document}